\documentclass[12pt]{article}

\usepackage[utf8]{inputenc}
\inputencoding{utf8}

\usepackage{booktabs}
\usepackage{algorithm, algorithmic}
\usepackage{blkarray}
\usepackage{amssymb}
\usepackage{amsmath}
\usepackage{amsthm}
\usepackage{url}
\usepackage{subcaption}
\usepackage{graphicx}
\usepackage{rotating}

\usepackage{tikz}
 \usetikzlibrary{arrows,plotmarks,decorations.markings,trees,shapes,pgfplots.groupplots,automata,circuits}
 \tikzset{dot/.style = {circle, fill, minimum size=#1,inner sep=0pt, outer sep=0pt, fill, circle},dot/.default = 6pt}
 \tikzset{dot2/.style = {circle, fill, color=black!40,minimum size=6pt,inner sep=0pt, outer sep=0pt, fill, circle}}
 \tikzstyle{a}=[->,>=stealth,dashed]
 \tikzstyle{a2}=[->,>=stealth]
 \tikzstyle{nodo}=[ellipse,draw=black!100,fill=black!0,line width=.7pt,minimum width=1.2cm,minimum height=0.8cm,text width=1.2cm,text centered]
 \tikzstyle{nodo2}=[ellipse,draw=black!100,fill=black!10,line width=.7pt,minimum width=1.2cm,minimum height=0.8cm,text width=1.2cm,text centered]
 \tikzstyle{nodo3}=[ellipse,draw=black!100,fill=black!30,line width=.7pt,minimum width=1.2cm,minimum height=0.8cm,text width=1.2cm,text centered]
 \tikzstyle{arco}=[draw=black!80,line width=.7pt, postaction={decorate}, decoration={markings,mark=at position 1.0 with {\arrow[ draw=black!80,line width=.7pt]{>}}}]

\newtheorem{dfn}{Definition}
\newtheorem{exe}{Example}

\usepackage{centernot}

\usepackage{soul}
\usepackage{multirow}

\newcommand{\todocite}[1]{\hl{[#1]}}	
\newcommand{\todoinline}[1]{\hl{TODO: #1}}	

\newcommand{\bm}[1]{\mathbf{#1}}
\newcommand{\bmU}{\bm{U}}

\newcommand{\bmX}{\bm{X}}
\newcommand{\bmY}{\bm{Y}}

\newcommand{\bmV}{\bm{V}}
\newcommand{\bmv}{\bm{v}}

\newcommand{\mcM}{\mathcal{M}}
\newcommand{\mcD}{\mathcal{D}}
\newcommand{\mcG}{\mathcal{G}}
\newcommand{\mcF}{\mathcal{F}}
\newcommand{\mcP}{\mathcal{P}}
\newcommand{\tP}{\tilde{P}}

\newcommand{\dop}{\mathrm{do}}

\usepackage{xargs}
\newcommand{\setFont}[1]{\ensuremath{\mathcal{#1}}}
\newcommandx{\pM}[1][1=M]{\ensuremath{\setFont{#1}}}  
\newcommand{\pMprime}{\pM[M']}
\newcommandx{\fM}[1][1=M]{\ensuremath{#1}}  
\newcommand{\solutionLetter}{\ensuremath{\setFont{S}}}
\newcommandx{\solution}[2][1=\pM, 2=\mcD]{\ensuremath{\solutionLetter_{#1, #2}}}  
\newcommand{\remove}[1]{}

\title{Causal computations in Semi Markovian Structural Causal Models using divide and conquer}
\author{Anna Rodum Bj{\o}ru \and Rafael Caba\~{n}as \and Helge Langseth \and Antonio Salmer\'{o}n}
\date{Sept 15th, 2025}
\begin{document}
\maketitle

\begin{abstract}

Recently, Bjøru et al.\cite{bjoru2024,bjoru2025ijar}  proposed a novel divide-and-conquer algorithm for bounding counterfactual probabilities in structural causal models (SCMs). 
They assumed that the SCMs were learned from purely observational data, leading to an imprecise characterization of the marginal distributions of exogenous variables.
Their method leveraged the canonical representation of structural equations to decompose a general SCM with high-cardinality exogenous variables into a set of sub-models with low-cardinality exogenous variables. 
These sub-models had precise marginals over the exogenous variables and therefore admitted efficient exact inference. 
The aggregated results were used to bound counterfactual probabilities in the original model. 
The approach was developed for Markovian models, where each exogenous variable affects only a single endogenous variable.
In this paper, we investigate extending the methodology to \textit{semi-Markovian} SCMs, where exogenous variables may influence multiple endogenous variables. Such models are capable of representing confounding relationships that Markovian models cannot.
We illustrate the challenges of this extension using a minimal example, which motivates a set of alternative solution strategies. These strategies are evaluated both theoretically and through a computational study.
\end{abstract}

\section{Introduction}
Structural causal models (SCMs) with discrete variables \cite{bareinboim2022pearl,pearl2009} are probabilistic graphical models (PGMs) designed for causal and counterfactual reasoning that enable reasoning about hypothetical scenarios. 
SCMs consist of endogenous (observable) and exogenous (usually latent) variables, with endogenous values determined from the exogenous ones through structural equations. 
Causal analysis is often described using \textit{Pearl's ladder} \cite{pearl2018}.
At the first rung are the observational queries (e.g., ``What if I \textit{see} this?''). Answering such queries based on passively observed data is the scope of standard statistical models, and the results are based on the calculation of conditional distributions. 
At the second rung we find interventional queries (e.g., ``What if I \textit{do} this?''). 
These are calculable by, for instance, causal Bayesian models \cite{pearl2009} and are represented by conditional distributions after interventions based on so-called \textit{do-calculus} \cite{pearl2009}. 
Finally, the last rung holds counterfactual queries that address hypothetical scenarios (e.g., ``What if we had done this instead of passively observing that?''). 
They are calculable by SCMs. However, when the SCM model is learned from data, the result of such queries are not necessarily precisely defined, and the answer to the query will then be  given as an imprecise probability, and represented through a (non-degenerate) interval. 
Unfortunately, the computational complexity of calculating these intervals exactly is overwhelming even for medium-sized models, and this has motivated research into computationally efficient approaches. 

The \textit{Divide and Conquer for Causal Computation} (DCCC) method~\cite{bjoru2024,bjoru2025ijar} can bound non-identifiable queries by, for each exogenous variable, solving a linear programming problem that characterizes all exogenous distributions consistent with the available data.  
Its main advantage over other approaches is that DCCC reduces the computational complexity of the problem by removing certain exogenous states, thereby transforming the model into a collection of simpler SCMs, each of which admits a unique solution to the corresponding linear program.
However, DCCC is designed for Markovian models, in which each exogenous variable is allowed to have only one endogenous child. In this work, we address the challenges of extending DCCC to semi-Markovian models, where  exogenous variables are allowed to have more than one endogenous child. Our main contributions are as follows:

\begin{itemize}
    \item We show how DCCC can be extended in a specific semi-Markovian topology to obtain the exact solution. In addition, we analyse how ignoring the confounding relations and applying DCCC in its Markovian form can be used to approximate the solution.
    \item We extend DCCC to accommodate multiple data sources (observational and experimental) and examine how the type of data influences the informativeness of the solution.
\end{itemize}

The remainder of this paper is organized as follows.  
Section~\ref{sec:background} introduces the fundamental definitions and notation related to SCMs.  
A summary of DCCC for Markovian models is provided in Section~\ref{sec:markovianmodels}.  
The main contributions are presented in the subsequent two sections:  
Section~\ref{sec:semimarkovian} describes the exact extension of DCCC to semi-Markovian models, incorporating both observational and experimental data;  
Section~\ref{sec:markovianapprox} presents approximate versions of DCCC obtained by omitting confounding relations.  
Section~\ref{sec:experiments} reports the experimental evaluation, and Section~\ref{sec:conclusions} concludes the paper and gives directions for future research.

\section{Background}\label{sec:background}

\subsection{Basic notation}
We use upper-case letters to denote random variables and lower-case for their possible values (or states), while $\Omega_V$ denotes the state space of variable $V$. 
Similarly, $\bmV =\{V_1, V_2,\ldots,V_n\}$ denotes a set of variables and $\bmv$ a joint state of its domain $\Omega_\bmV = \displaystyle\times_{V \in \bmV} \Omega_V$. 
Variables are omitted from assignments when their context is clear, and we may for instance simply denote $P(V=v)$  as $P(v)$. 
We will sometimes abuse notation by writing that, e.g., $P(X|Y)=g_1(X,Y)$ as a shorthand to denote that the relationship $P(X=x|Y=y)=g_1(x,y)$ holds for all $x\in\Omega_X$ and $y\in\Omega_Y$. 
Similarly, we may also write $P(X|\dop(Y))=g_2(X,Y)$ to describe a general relationship between $P(X=x|\dop(Y=y))$ and $g_2(x,y)$ that will hold for  all $x\in\Omega_X$ and $y\in\Omega_Y$.
Finally, $\tP$ is used to denote empirical distributions.

\subsection{Structural causal models}
Structural causal models (SCMs) \cite{pearl2009} are a class of PGMs used for causal and counterfactual reasoning, consisting of two types of nodes: \emph{endogenous} nodes, which represent the internal variables of the modelled problem, and \emph{exogenous} nodes, which represent factors outside the model. Formally, an SCM $\mcM$ is a 4-tuple $\langle \bmU, \bmV,\mcF, \mcP \rangle$, where~\cite{bareinboim2022pearl}

	\begin{itemize}
		\item $\bmU$ is a set of exogenous variables that are determined by factors outside the model,
		\item $\bmV$ is a set of variables $\{V_1, V_2,\ldots,V_n\}$, called endogenous, that are determined by other (exogenous and endogenous) variables in the model,  i.e. by variables in $\bmU \cup \bmV$, 
		\item $\mcF$ is a set of functions $\{f_{V_1}, f_{V_2},..., f_{V_n}\}$ called \textit{structural equations} (SE), such that each of them is a function 
        $f_{V_i}:\Omega_{\mathrm{Pa}_{V_i}} \to \Omega_{V_i}$, where $\mathrm{Pa}_{V_i}$ are the  variables directly determining $V_i$, including the exogenous parent
        $U_i\subseteq \bmU$,
		\item $\mcP$ is a set of probability distributions $P(U)$ for each $U\in\bmU$.
	\end{itemize}

The structural equations $\mcF$ define a directed acyclic graph (DAG) $\mcG$, referred to as the \emph{causal graph} of the model, where the nodes correspond to the variables in $\bmU \cup \bmV$. The domains of both the endogenous and exogenous variables are assumed to be finite and discrete. A discrete exogenous variable, taking values on a finite set of unobserved states, is enough to represent all counterfactual distributions~\cite{zhang2022partial}.

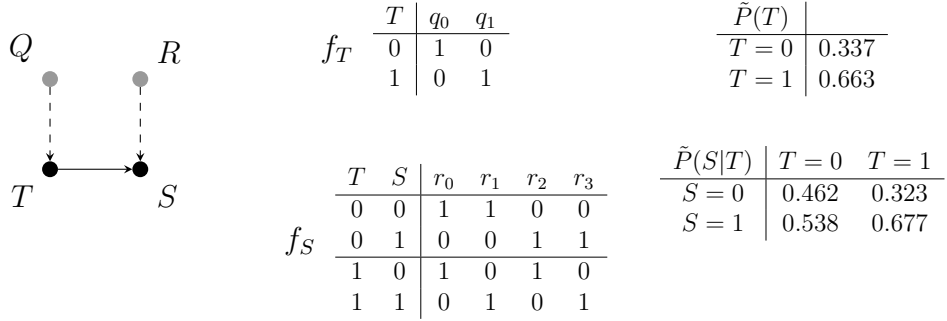
\begin{figure}[htp!]
	\centering
		\begin{tikzpicture}[scale=0.8]
			\node[dot,label=below left:{$T$}] (T)  at (-1.5,0) {};
			\node[dot,label=below right:{$S$}] (S)  at (0,0) {};
			\node[dot2,label=above left:{$Q$}] (V)  at (-1.5,1.5) {};
			\node[dot2,label=above right:{$R$}] (U)  at (0,1.5) {};
			\draw[a2] (T) -- (S);
			\draw[a] (V) -- (T);
			\draw[a] (U) -- (S);

            \node[scale=0.8,label=left:{$f_T$}] at (5, 2.0){
            \begin{tabular}{c|cc}
            $T$ & $q_0$ & $q_1$ \\
            \hline 
            0 & 1 & 0 \\
            1 & 0 & 1
            \end{tabular}
            };

            \node[scale=0.8,label=left:{$f_S$}] at (5.5, -1.2){
            \begin{tabular}{cc|cccc}
            $T$ & $S$ & $r_0$ & $r_1$ & $r_2$ & $r_3$\\
            \hline 
            0 & 0 & 1 & 1 & 0 & 0 \\
            0 & 1 & 0 & 0 & 1 & 1 \\
            \hline
            1 & 0 & 1 & 0 & 1 & 0 \\
            1 & 1 & 0 & 1 & 0 & 1 \\
            
            \end{tabular}
            };

            \node[scale=0.8] at (11, 2.0){
            \begin{tabular}{c|c}
             $\tilde{P}(T)$ & \\ \hline 
             $T=0$ & 0.337 \\
             $T=1$ & 0.663 
            \end{tabular}
            };

            \node[scale=0.8] at (11, -0.4){
            \begin{tabular}{c|cc}
             $\tilde{P}(S|T)$ & $T=0$ & $T=1$ \\
            \hline 
             $S=0$& 0.462 & 0.323 \\
             $S=1$ & 0.538 & 0.677
            \end{tabular}
            };
		\end{tikzpicture}
	\caption{Elements of an SCM: (left) causal graph, (center) structural equations and (right) empirical distribution computed from the data.
 }\label{fig:example_scm}
\end{figure}

When the distributions for the exogenous variables are not known, we refer to the model as a \textit{partially} specified SCM, denoted by a calligraphic letter, such as \pM. In contrast, if all such distributions are provided, the model is considered fully specified and is denoted as, e.g., \fM. 
Figure~\ref{fig:example_scm}, shows an example of an SCM~\cite{mueller2021causes}. The causal graph on the left includes the endogenous variables $\bmV = \{T, S\}$, representing the \textit{treatment} and \textit{survival}, respectively. The goal is to analyze whether receiving treatment ($T=1$) contributes to survival ($S=1$). 
Meanwhile, $\bmU = \{Q, R\}$ represents the set of exogenous variables, which are assumed to be root nodes, with endogenous variables as their children. The SEs $f_T(Q)$ and $f_S(T,R)$ are represented
as deterministic CPTs of the form $P(T|Q)$ and $P(S|T,R)$, characterized by containing only ones and zeros. 
If not provided from expert knowledge, SEs can be automatically inferred from the causal graph, without any loss of generality, via a \textit{canonical specification}. 
This is the case of this example, where the states of an exogenous variable will then represent all possible deterministic mechanisms between its child and the child's endogenous parents.

The SCM in Figure~\ref{fig:example_scm} assumes \textit{unconfoundedness}, meaning that there are no unobserved confounders, i.e., no unobserved common causes of two or more endogenous variables. According to the classification introduced in~\cite{avin2005identifiability}, an SCM with such a structure is referred to as \textit{Markovian}, where each exogenous variable influences exactly one endogenous variable. In contrast, when an exogenous variable is allowed to have multiple endogenous children, the model is termed \textit{semi-Markovian}\footnote{Some authors adopt a more restrictive definition of semi-Markovian models, limiting each exogenous variable to have no more than two children~\cite{huang2006identifiability}.}. It is important to note that each endogenous variable must have exactly one exogenous parent in both Markovian and semi-Markovian models. 
In the Markovian case, the cardinality of each exogenous variable $U$ is at most $|\Omega_Y|^{|\Omega_{\bmX}|}$, where $Y$ denotes its only endogenous child and $\bmX$ the set of endogenous parents of $Y$.

When performing inference with an SCM $\pM$ and a dataset $\mcD$, one typically has access only to observations of the endogenous variables. In the case of the model under consideration, the dataset $\mcD$ allows us to compute the empirical distributions $\tP(T)$ and $\tP(S \mid T)$, whose values are depicted in Figure~\ref{fig:example_scm} (right). The problem of inference in a partial SCM given a dataset can be framed as estimating the collection of fully-specified SCMs that remain consistent with the observed data. 



\section{DCCC for Markovian models}
\label{sec:markovianmodels}

\subsection{Imprecise characterisation}

The task of doing inference in a partial SCM given a dataset involves estimating a set of fully-specified SCMs that are compatible with the data~\cite{bjoru2024}. SCMs can be represented as \textit{credal networks}~\cite{zaffalon2020structural,zaffalon2024efficient} which are Bayesian networks with imprecise conditional distributions~\cite{cozman2000credal}. 
In credal networks, each node is associated with a set of probability mass functions known as a \textit{credal set}. 
When transforming SCMs into credal networks, the endogenous observations impose linear constraints on the probabilities of the exogenous variables, so that a separate credal set for each exogenous variable can be obtained, called the \textit{solution set}. 
The solution set is the convex set of all the distributions over an exogenous variable $U$ leading to the same distribution over the endogenous children after marginalizing out the exogenous parents. This convex set can be defined as
\begin{equation}\label{eq:exo_credal_set}
    \setFont{K}(U):= \left\{P(U):\sum_{u\in\Omega_U} P(u)\cdot P(Y|\bmX,u) = \tilde{P}(Y|\bmX)\right\},
\end{equation}
where $Y$ is the only child of $U$, $\bmX$ the set of endogenous parents of $Y$ and $\tilde{P}(Y|\bmX)$ is the empirical distribution computed from $\mcD$. 
In our framework, the objective is to identify all such distributions.  
We define the solution set for a partially-specified SCM $\pM$ given a dataset $\mcD$, denoted by \solution, as the collection of all fully-specified SCMs satisfying $P(U) \in \setFont{K}(U)$ for every $U \in \bmU$.  
In other words, each SCM $M\in \solution$ constitutes a valid solution for the given dataset $\mcD$ and partially-specified SCM $\pM$.  
Once \solution\ has been identified, it can be employed to bound any causal or counterfactual query by evaluating the query within each model in the set and subsequently aggregating the results.

\subsection{Example Markovian SCM}
\label{sec:markovianexamplescm}
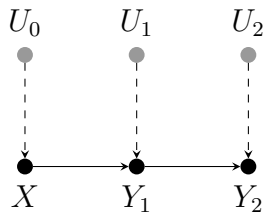
\begin{figure}[htp!]
	\centering
		\begin{tikzpicture}[scale=0.99]
			\node[dot,label=below:{$X$}] (X)  at (-1.5,0) {};
			\node[dot,label=below:{$Y_1$}] (Y1)  at (0,0) {};		
            \node[dot,label=below:{$Y_2$}] (Y2)  at (1.5,0) {};
			\node[dot2,label=above :{$U_0$}] (V)  at (-1.5,1.5) {};
			\node[dot2,label=above :{$U_1$}] (U1)  at (0,1.5) {};

            \node[dot2,label=above :{$U_2$}] (U2)  at (1.5,1.5) {};
			\draw[a2] (X) -- (Y1);
            \draw[a2] (Y1) -- (Y2);
			\draw[a] (V) -- (X);
			\draw[a] (U1) -- (Y1);
            \draw[a] (U2) -- (Y2);

		\end{tikzpicture}
	\caption{Graph of the example Markovian model.}
    \label{fig:examplemodel_markovian}
\end{figure}

In order to present our approach, we start by describing the DCCC algorithm for Markovian models. For concreteness we tie the discussions to a specific example model, namely the model given in Figure~\ref{fig:examplemodel_markovian}. 
The three endogenous variables $X$, $Y_1$ and $Y_2$ all have binary domains. 
There are three associated exogenous variables $U_0,U_1, U_2$. The variables $X, Y_1, Y_2, U_0, U_1, U_2$ along with a set of structural functions $\mcF = \{f_0, f_1, f_2\}$, where
\begin{align*}
    X &= f_0(U_0) ,  \\
    Y_1 &= f_1(X, U_1) = f_{1, U_1}(X) , \\
    Y_2 &= f_2(Y_1, U_2) = f_{2, U_2}(Y_1) ,
\end{align*}
make up a partially specified SCM $\pM$. 
Adopting the most general specification of $\pM$ ensures that, given a dataset $\mcD$, any fully specified model $M$ compatible with $\mcD$ will be in the solution space 
$\solution$. 

A canonical specification for any exogenous variable $U$ with endogenous child $Y$ in a Markovian SCM, defines its domain $\Omega_U$ such that each possible deterministic structural function from $\bmX = \mathrm{Pa}_Y \setminus \{U\}$ to $Y$ is indexed by a distinct $u_i \in \Omega_U$. Thus, $|\Omega_U| = |\Omega_Y|^{|\Omega_{\bmX}|}$ for a canonical model. 
Given the example SCM with structure shown in Figure \ref{fig:examplemodel_markovian}, defining a canonical $\Omega_{U_0}$ is trivial, with $X=f_0({U_0})={U_0}$ being the most general specification, from which $\Omega_{U_0} = \Omega_X$ and  $P({U_0}) = \tP(X)$ follows. Moving on to $Y_1$, which a single endogenous parent, we find that $|\Omega_{U_1}| = 4 $, with $f_{1,U_1}(X) \in \{0, X, \overline{X}, 1\}$. The canonical specification for $U_1$ is given in Table~\ref{tab:U1}, which corresponds with the relation $Y_1=f_1(X, U_1)$. 
\begin{table}[ht]
\centering
\scalebox{0.9}{
\begin{tabular}{ |c|c||c|c|c|c||c| } 
 \hline
 \multirow{3}{1.6em}{$X$} & \multirow{3}{1.6em}{$Y_1$}& \multicolumn{4}{|c||}{$U_1$} & \multirow{3}{4em}{$\tP(Y_1 | X)$}\\ 
 \cline{3-6}
  $ $ & $ $ & $u_{1,0}$ & $u_{1,1}$ & $u_{1,2}$ & $u_{1,3}$ & $ $\\
  \cline{3-6}
  & &\scriptsize$f_1(X) = 0$ & \scriptsize$f_1(X) = X$ & \scriptsize$f_1(X) = \overline{X}$ & \scriptsize$f_1(X) = 1$ & $ $ \phantom{$X^{X^X}$}\\
 \hline
 \hline
 0 & 0 & 1 & 1 & 0 & 0 &  \scriptsize $\tP(Y_1=0|X=0) = P(u_{1,0})+P(u_{1,1})$\\
  0 & 1 & 0 & 0 & 1 & 1 &  \scriptsize $\tP(Y_1=1|X=0) = P(u_{1,2})+P(u_{1,3})$\\
 \hline
 1 & 0 & 1 & 0 & 1 & 0 &  \scriptsize $\tP(Y_1=0|X=1) = P(u_{1,0})+P(u_{1,2})$\\
 1 & 1 & 0 & 1 & 0 & 1 &  \scriptsize $\tP(Y_1=1|X=1) = P(u_{1,1})+P(u_{1,3})$\\
 \hline
\end{tabular}
}\vspace{5pt}
\caption{A canonical specification of the domain $\Omega_{U_1}$ for variable $U_1$ in Figure \ref{fig:examplemodel_markovian}. Each $u_{1,i} \in \Omega_{U_1}$ corresponds to a distinct deterministic function $f_1: \Omega_X \rightarrow \Omega_{Y_1}$. 
For each pair of values $u_{1,i}, x_j$, the table details the (deterministic) probabilities $P(Y_1|X=x_j, U_1=u_{1,i})$ that are a consequence of the defined structural function $f_{1,U_1}$. Given these distributions, the rightmost column deconstructs the equations constraining the credal set $\setFont{K}(U_1)$ (Equation \ref{eq:csetU1}) for each pair of values $X=x_j, Y_1=y_k$. 
}
\label{tab:U1}
\end{table}
The four columns of the table describe the four structural equations that are possible. 
Each row gives one configuration over the endogenous data, the first row does for instance consider the configuration $(X=0, Y_1=0)$.
The $i$'th cell of a row with configuration $(x_j, y_j)$ is defined as $I(f_1(x_j, u_{1,i}) = y_{j})$, 
where $I(\phi)$ is the indicator-function that returns 1 if $\phi$ is true and 0 otherwise. 
Sometimes we will also represent this relationship as a (deterministic) probability distribution, and define
$P(Y_1=y_j|X=x_j, U_1=u_i) = I(f_1(x_j, u_{1,i}) = y_{j})$.

The situation is analogous for $Y_2$, which also has a single endogenous parent.  
Consequently, we have $|\Omega_{U_2}| = 4$ and $f_{2,U_2}(Y_1) \in \{0,\, Y_1,\, \overline{Y_1},\, 1\}$.  
The corresponding canonical specification for $U_2$ is presented in Table~\ref{tab:U2}.

\begin{table}[ht]
\centering
\scalebox{0.85}{
\begin{tabular}{ |c|c||c|c|c|c||c| } 
 \hline
 \multirow{3}{1.6em}{$Y_1$} & \multirow{3}{1.6em}{$Y_2$}& \multicolumn{4}{|c||}{$U_2$} & \multirow{3}{4em}{$\tP(Y_2 | Y_1)$}\\ 
 \cline{3-6}
  $ $ & $ $ & $u_{2,0}$ & $u_{2,1}$ & $u_{2,2}$ & $u_{2,3}$ & $ $\\
 \cline{3-6}
   & &\scriptsize$f_2(Y_1) = 0$ & \scriptsize$f_2(Y_1) = Y_1$ & \scriptsize$f_2(Y_1) = \overline{Y_1}$ & \scriptsize$f_2(Y_1) = 1$ & $ $ \phantom{$X^{X^X}$}\\
 \hline
 \hline
 0 & 0 & 1 & 1 & 0 & 0 &  \scriptsize $\tP(Y_2=0|Y_1=0) = P(u_{2,0})+P(u_{2,1})$\\
  0 & 1 & 0 & 0 & 1 & 1 &  \scriptsize $\tP(Y_2=1|Y_1=0) = P(u_{2,2})+P(u_{2,3})$\\
 \hline
 1 & 0 & 1 & 0 & 1 & 0 &  \scriptsize $\tP(Y_2=0|Y_1=1) = P(u_{2,0})+P(u_{2,2})$\\
 1 & 1 & 0 & 1 & 0 & 1 &  \scriptsize $\tP(Y_2=1|Y_1=1) = P(u_{2,1})+P(u_{2,3})$\\
 \hline
\end{tabular}
}
\caption{A canonical specification of domain $\Omega_{U_2}$ for $U_2$ in Fig. \ref{fig:examplemodel_markovian}. The table is read analogously to Table \ref{tab:U1}, with the rightmost column itemising the constraints of credal set $\setFont{K}(U_2)$ (Equation \ref{eq:csetU2}).
}
\label{tab:U2}
\end{table}

The credal set definition given in Equation~\eqref{eq:exo_credal_set} can now be defined for the model in Figure~\ref{fig:examplemodel_markovian} as follows:
\begin{equation}\label{eq:csetU1}
 \begin{aligned}
    \setFont{K}(U_1):= \left\{P(U_1): \sum_{u\in\Omega_{U_1}} [P(u)\cdot P(Y_1|X,u)] = \tP(Y_1|X) \right\}
\end{aligned}   
\end{equation}
\begin{equation}\label{eq:csetU2}
 \begin{aligned}
 \setFont{K}(U_2):= \left\{P(U_2):
     \sum_{u\in\Omega_{U_2}} [P(u)\cdot P(Y_2|Y_1,u)] = \tP(Y_2|Y_1) \right\}
\end{aligned}   
\end{equation}
The linear constraints corresponding to $\setFont{K}(U_1)$ and $\setFont{K}(U_2)$ are reported in the rightmost columns of Tables~\ref{tab:U1} and~\ref{tab:U2}, respectively.

\subsection{Overview of DCCC}

The DCCC method~\cite{bjoru2024,bjoru2025ijar} consists of transforming a complex SCM into a set of simpler ones, each with exogenous variables of smaller cardinality. 
The transformation involves the removal of some states from the exogenous domains.
This operation is known as a \textit{reduction} (also called \textit{branching} in the optimization literature), which can be defined as follows:

\begin{dfn}[Reduction operator] Let $\pM$ be a partially-specified SCM whose set of exogenous variables is $\bmU$ and let $u \in \Omega_U$ with $U \in \bmU$. Then the reduction operation, denoted $R(\pM, u)$, produces a new partially-specified SCM  $\pMprime$ by removing assignments from $\mcF$ and $\mcP$ in $\pM$ that are consistent with $u$. 
\end{dfn}

Reduction is applied to various states of the different exogenous variables, i.e., to a set defined as $\setFont{A}_\bmU := \left\{u^{(i)}\right\}_{i=1}^{m}$ s.t. $u^{(i)} \in \Omega_U$ and $U\in\bmU$.   
For simplicity we can recursively define this as $R(\pM, \setFont{A}_\bmU) = R\left(R(\pM, u^{(1)}), \setFont{A}_\bmU\setminus\{u^{(1)}\}\right)$. The reduction operation $R(\pM, u)$ is equivalent to imposing the constraint  $P(u) = 0$ on $\pM$. Thus, when looking for the solution set of a reduced SCM, we can equivalently look
for all the solutions from the original solution set that are consistent with that constraint.


\begin{exe}

Consider the problem of identifying all possible exogenous distributions for $P(R)$ and $P(Q)$ in the model $\mcM$ depicted in Figure~\ref{fig:example_scm}, which are consistent with a dataset following the distribution $\tilde{P}(S,T) = \tilde{P}(S \mid T) \cdot \tilde{P}(T)$. In this setting, all admissible solutions for $P(R)$ must satisfy the following system of linear constraints.

\begin{align*}
    P(r_0) + P(r_1) &= 0.462, \\
    P(r_2) + P(r_3) &= 0.538, \\
    P(r_0) + P(r_2) &= 0.323, \\
    P(r_1) + P(r_3) &= 0.677.
\end{align*}

\noindent This system has an infinite set of solutions. To derive a unique solution, the DCCC algorithm applies the reduction operator recursively. For example, applying $R(\mathcal{M}, r_2)$ yields:

\begin{align*}
    P(r_0) + P(r_1) &= 0.462, \\
    P(r_3) &= 0.538, \\
    P(r_0)  &= 0.323, \\
    P(r_1) + P(r_3) &= 0.677.
\end{align*}

Similarly, applying $R(\mathcal{M}, r_0)$ produces:

\begin{align*}
    P(r_1) &= 0.462, \\
    P(r_2) + P(r_3) &= 0.538, \\
    P(r_2) &= 0.323, \\
    P(r_1) + P(r_3) &= 0.677.
\end{align*}

\noindent The corresponding solutions for these systems are, respectively:

\[
P_1(R)=
			\begin{blockarray}{cccc}
				\color{gray}{r_0} &  \color{gray}{r_1} & \color{gray}{r_2} &  \color{gray}{r_3}  \\
				\begin{block}{[cccc]}
					0.323 & 0.139 & 0 & 0.538\\
				\end{block}
			\end{blockarray} ,
\phantom{=}
P_2(R)=
			\begin{blockarray}{cccc}
				\color{gray}{r_0} &  \color{gray}{r_1} & \color{gray}{r_2} &  \color{gray}{r_3}  \\
				\begin{block}{[cccc]}
					0 & 0.462 & 0.323 & 0.215\\
				\end{block}
			\end{blockarray} .
\]
\noindent By contrast, $R(\mathcal{M}, r_1)$ and $R(\mathcal{M}, r_3)$ lead to unsolvable equation systems. $P_1(R)$ and $P_2(R)$ represent the extreme points of the convex set containing all feasible solutions. Equivalently, any solution can be expressed as a convex combination of these points.

\end{exe}

Once the solution set is identified, any query in the model can be bounded as follows. For each extreme point, a fully specified SCM is built and used to compute the query. The lower and upper bounds are then obtained from the minimum and maximum values across all such computations.

\subsection{Solution search}
\label{sec:solutionseach_MDCCC}


The DCCC method for Markovian models is applied to each exogenous variable $U$ in a model. Each variable $U$ has a single endogenous child $Y$, with endogenous parents $\bmX = \mathrm{Pa}_Y \setminus \{U\}$, all variables assumed discrete. With $p=|\Omega_{\bmX}|$ and  $q = |\Omega_Y|$, the canonical domain size of $U$ is $|\Omega_U| = q^{p}$, and the linear constraints defining the solution space are $\sum_{u\in\Omega_U} P(u)\cdot P(Y|\bmX, u) = \tP(Y|\bmX)$. There are $q\cdot p$ such constraints, of which $p\cdot (q-1) +1$ are independent under $\sum_{u\in\Omega_U} P(u) = 1$. Extreme points in $ \setFont{K}(U):= \{P(U): \sum_{u\in\Omega_{U}} [P(u)\cdot P(Y|\bmX,u)] = \tP(Y|\bmX) \}$ are thus distributions $P(U)$ for which at most ${p\cdot(q-1) +1}$ of the states $u \in \Omega_U$ have non-zero probability. See~\cite{bjoru2025ijar} for details and proof.

Extreme points in $\setFont{K}(U)$ therefore correspond to reductions $R$ of the original model $\mcM$ such that the image $U'$ of $U$ under $R$ has domain size at most $|\Omega_{U'}| = p\cdot(q-1)+1$. The space of possible such reductions is of size  
$\binom{q^p}{p\cdot(q-1) +1}$. The exhaustive DCCC extreme distribution search considers all $\binom{q^p}{p\cdot(q-1) +1}$ possible selections of $p\cdot(q-1) +1$ states from the original domain $\Omega_U$, sets the probability of the states in the complement set to 0, and solves for the unique solution of the resulting linear system. If this solution is a probability distribution such that $P(u) \geq 0, \forall u \in \Omega_U$, $\sum_{u\in\Omega_U} P(u) = 1$, it is retrieved as an extreme point of $\setFont{K}(U)$. Once all $\binom{q^p}{p\cdot(q-1) +1}$ possibilities are explored, a complete collection of extreme points is retrieved and exact query bounds may be calculated. 

The Markovian DCCC algorithm requires only $p$ and $q$ as inputs, and relies on a structured enumeration of the canonical domain of exogenous variable $U$ in order to efficiently retrieve extreme points. The states are ordered such that for a state $u_i$, the base-$q$ encoding of $i$ to $p$ digits reveals for which constraints $\sum_{u\in\Omega_{U}} [P(u)\cdot P(Y|\bmX,u)] = \tP(Y|\bmX)$, $P(u_i)$ will appear on the left hand side. That is, for which value combinations $y_k \in \Omega_Y, \bm{x}_j \in \Omega_\bmX$ it is the case that $P(Y=y_k|\bmX=\boldsymbol{x}_j,u_i) = 1$. Namely, for a fixed order of states $\bm{x}_j \in \Omega_\bmX$, $P(Y=y_k|\bmX=\boldsymbol{x}_j,u_i) = 1$ if and only if position $j$ in the base-$q$ encoding of $i$ equals $y_k$. 

With this implicit representation of the complete linear system of $\setFont{K}(U)$, the reduced linear systems corresponding to each $\setFont{K}(U')$ is generated directly from the collection of states selected, rather than requiring an explicit representation of the complete system prior to reduction. This representation trick is further exploited in the heuristic version of the DCCC extreme distribution search, where possible model reductions are selected more strategically than via exhaustive iteration. One heuristic is based on ensuring the selection of states is such that each constraint has at least one non-zero $P(u_i)$ component part of its equation, which excludes a subset of the non-viable state selections from the search. A second heuristic restricts the number of non-zero $P(u_i)$ in the constrains summing to the lowest observed probabilities, in order to focus the search towards solutions more likely to correspond to probability distributions. As $q^p$ grows with increased model sizes, these heuristics help speed up the extreme point search and allows calculation of approximate query bounds when the exhaustive search is no longer feasible. For details on the techniques of the DCCC solution search, see \cite{bjoru2025ijar}. 

\section{DCCC for semi-Markovian models}
\label{sec:semimarkovian}

We will now see how to generalize DCCC to semi-Markovian models. 
We start with a short review of the state of the art in semi-Markovian models in Section \ref{sec:sota}, before describing our DCCC extension. This is divided into two parts: 
We will first look at how to handle purely observational data in Section \ref{sec:semimarkovianobs}, thereafter consider experimental data in Section \ref{sec:semimarkovianexp}.

\subsection{Related work}
\label{sec:sota}

Several methods have been proposed for bounding non-identifiable queries in semi-Markovian models. This problem was first addressed by systematically deriving constraints on a causal query, although this approach suffers from exponential computational complexity~\cite{kang2012inequality}. An alternative approach formulates the bounding problem as a symbolic linear program, allowing for the exact computation of bounds~\cite{sachs2020}. More recently, an entropy minimization based optimization method to approximate bounds on causal effects was proposed~\cite{jiang2023approximate}. While the previous approaches are capable of handling semi-Markovian models, they are generally limited to computing observational queries.  Some methods can also compute counterfactual queries, also in semi-Markovian models. 
One approach involves transforming the SCM into a credal network~\cite{cozman2000credal}, which requires solving a linear programming problem~\cite{zaffalon2020structural} for each exogenous variable. However, this method may become infeasible when the number of exogenous variables is large, due to the exponential growth in the number of possible configurations. To address this, approximate bounds for any non-identifiable query can be obtained using the EMCC method~\cite{zaffalon2024efficient}, which repeatedly runs the Expectation-Maximization (EM) algorithm~\cite{koller2009probabilistic} to estimate precise specifications of the exogenous distributions. Once these distributions are obtained, queries can be computed separately and aggregated to approximate the  bounds. Nevertheless, each individual EM run may require a very large number of iterations to reach a sufficiently low error. A related method instead generates the exogenous distributions using Gibbs sampling techniques~\cite{zhang2021}.

\subsection{Observational data}
\label{sec:semimarkovianobs}

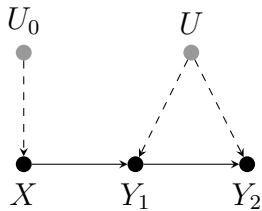
\begin{figure}[htp!]
	\centering
		\begin{tikzpicture}[scale=0.99]
			\node[dot,label=below:{$X$}] (X)  at (-1.5,0) {};
			\node[dot,label=below:{$Y_1$}] (Y1)  at (0,0) {};		
            \node[dot,label=below:{$Y_2$}] (Y2)  at (1.5,0) {};
			\node[dot2,label=above :{$U_0$}] (V)  at (-1.5,1.5) {};
			\node[dot2,label=above :{$U$}] (U)  at (0.75,1.5) {};
			\draw[a2] (X) -- (Y1);
            \draw[a2] (Y1) -- (Y2);
			\draw[a] (V) -- (X);
			\draw[a] (U) -- (Y1);
            \draw[a] (U) -- (Y2);

		\end{tikzpicture}
	\caption{Graph of the example model for semi Markovian discussion.}
    \label{fig:modelexample}
\end{figure}

We will now consider how to adapt DCCC to the more general case of \textit{semi-Markovian} models. 
Again, we will do so using an example model, as we are adapting the Markovian example model from Section \ref{sec:markovianexamplescm}. 
Also the semi-Markovian model has three endogenous variables, $X, Y_1, Y_2$, all with binary domains. 
As before, $X$ has the exogenous parent $U_0$, but the variables $Y_1$ and $Y_2$ now share a common exogenous parent $U$ introducing a confounder between these endogenous variables. 
The updated structure is shown in Figure \ref{fig:modelexample}. 

The variables $X, Y_1, Y_2, U_0, U$ along with structural functions
\begin{align*}
    X &= f_0(U_0) ,  \\
    Y_1  &= f_1(X, U) = f_{1, U}(X) , \\
    Y_2 &= f_2(Y_1, U) = f_{2, U}(Y_1) ,
\end{align*}
make up a partially specified semi-Markovian SCM $\pM$.

\begin{center}
\begin{sidewaystable}[htp]
\scriptsize
\centering
\begin{tabular}{ | p{17pt} | p{17pt} |p{17pt}||p{9pt}|p{9pt}|p{9pt}|p{9pt}|p{9pt}|p{9pt}|p{9pt}|p{9pt}|p{9pt}|p{11pt}|p{11pt}|p{11pt}|p{11pt}|p{11pt}|p{11pt}|p{11pt}||c| } 
 \hline
  \multicolumn{3}{|c||}{$ $}& \multicolumn{16}{|c||}{$U$} & \\ 
 \cline{4-19}
    \multicolumn{3}{|c||}{$ $} & $u_0$ & $u_1$ & $u_2$ & $u_3$ & $u_4$ & $u_5$ & $u_6$ & $u_7$ & $u_8$ & $u_9$ & $u_{10}$ & $u_{11}$ & $u_{12}$ & $u_{13}$ & $u_{14}$ & $u_{15}$ & \\
\cline{1-19}
 \multicolumn{3}{|c||}{$f_{1,U}(X) = $} & $0$ & $0$ & $0$ & $0$  & $X$ & $X$ & $X$ & $X$ & $\overline{X}$ & $\overline{X}$ & $\overline{X}$  & $\overline{X}$  & $1$  & $1$ & $1$ & $1$ & $ $\phantom{$X^{X^{X^X}}$}\\
\cline{1-19}
    \multicolumn{3}{|c||}{$f_{2,U}(Y_1)=$} & 0 & $Y_1$ & $\overline{Y_1}$ & 1 & 0 & $Y_1$ & $\overline{Y_1}$ & 1  & 0 & $Y_1$ & $\overline{Y_1}$ & 1 &0 & $Y_1$ & $\overline{Y_1}$ & 1  & $ $\phantom{$X^{X^{X^X}}$}  \\
 \cline{1-19}
 \multicolumn{3}{|c||}{$f_{2,U}\circ f_{1,U}(X) = $} & $0$ & $0$ & $1$ & $1$ & $0$ & $X$ & $\overline{X}$ & $1$ & $0$ & $\overline{X}$ & $X$ & $1$ & $0$ & $1$ & $0$ &  $1$&$ $\phantom{$X^{X^{X^X}}$}  \\
    
 \hline
 \hline
 \multirow{2}{7pt}{$X$} & \multirow{2}{7pt}{$Y_1$}&  \multirow{2}{7pt}{$Y_2$}&\multicolumn{16}{c||}{$ $} & \multirow{2}{4em}{$\tP(Y_1, Y_2 | X)$}\\ 
  & & & \multicolumn{16}{c||}{$ $} & \\
  \hline
  \hline
  0 & 0 & 0 & 1 & 1 & 0 & 0 & 1 & 1 & 0 & 0 & 0 & 0 & 0 & 0 & 0 & 0 & 0 & 0 & \tiny $\tP(Y_1=0, Y_2=0| X=0)$\\
  0 & 0 & 1 & 0 & 0 & 1 & 1 & 0 & 0 & 1 & 1 & 0 & 0 & 0 & 0 & 0 & 0 & 0 & 0  & \tiny $\tP(Y_1=0, Y2=1| X=0)$\\
  0 & 1 & 0 & 0 & 0 & 0 & 0 & 0 & 0 & 0 & 0 & 1 & 0 & 1 & 0 & 1 & 0 & 1 & 0  & \tiny $\tP(Y_1=1, Y_2=0| X=0)$\\
  0 & 1 & 1 & 0 & 0 & 0 & 0 & 0 & 0 & 0 & 0 & 0 & 1 & 0 & 1 & 0 & 1 & 0 & 1  & \tiny $\tP(Y_1=1, Y_2=1| X=0)$\\
  \hline
  1 & 0 & 0 & 1 & 1 & 0 & 0 & 0 & 0 & 0 & 0 & 1 & 1 & 0 & 0 &  0 & 0 & 0 & 0 & \tiny $\tP(Y_1=0, Y_2=0| X=1)$\\
  1 & 0 & 1 & 0 & 0 & 1 & 1 & 0 & 0 & 0 & 0 & 0 & 0 & 1 & 1 & 0 & 0 & 0 & 0  & \tiny $\tP(Y_1=0, Y2=1| X=1)$\\
  1 & 1 & 0 & 0 & 0 & 0 & 0 & 1 & 0 &1 & 0 & 0 & 0 & 0 & 0 & 1 & 0 & 1 & 0  & \tiny $\tP(Y_1=1, Y_2=0| X=1)$\\
  1 & 1 & 1 & 0 & 0 & 0 & 0 & 0 & 1 & 0 & 1 & 0 & 0 & 0 & 0 & 0 & 1 & 0 & 1  & \tiny $\tP(Y_1=1, Y_2=1| X=1)$\\
 \hline

\end{tabular}
\caption{A canonical representation of $\Omega_U$ in the semi-Markovian model of \protect{Figure \ref{fig:modelexample}}, and leading to the credal set of Equation \eqref{eq:cssemimarkoviandef}.
The table is read analogously to Tables \ref{tab:U1} and \ref{tab:U2}.
}
\label{tab:semimarkovianobsmatrix}
\end{sidewaystable}
\end{center}

The canonical specification of this model should ensure that $\Omega_U$ contains states indexing each possible distinct combination of functions $f_{1,U}: \Omega_X \rightarrow \Omega_{Y_1}$ and $f_{2,U}: \Omega_{Y_1} \rightarrow \Omega_{Y_2}$. 
As we saw in Section \ref{sec:markovianexamplescm}, both $f_{1, U}$ and $f_{2, U}$ can be any of four different structural equations, hence the canonical representation for $U$ has size $|\Omega_{U}| = |\Omega_{Y_1}|^{|\Omega_{X}|}\cdot|\Omega_{Y_2}|^{|\Omega_{Y_1}|} = 16$. 
By directly extending the configurations for $f_1$ and $f_2$ introduced in Tables \ref{tab:U1} and \ref{tab:U2}, respectively, we introduce their combination $f_{2,U}\circ f_{1,U}(X) = f_{2,U}\left( f_{1,U}(X)\right)$ to signify how $U$ simultaneously determines the relationship from $X$ to $Y_1$ and from $Y_1$ to $Y_2$ in Table \ref{tab:semimarkovianobsmatrix}. 
Each column in the table describes the behaviour for one value $u_j$ of $U$, and presents a unique combination of structural equations $f_1$ and $f_2$ together with the combination $f_{2,U}\circ f_{1,U}(X) = f_{2,U}\left(f_{1,U}(X)\right)$. 
Consider for instance the column $u_{14}$, corresponding to $f_{1, u_{14}}(X)=1$, $f_{2, u_{14}}(Y_1)=\overline{Y_1}$. 
In this case $f_{2,u_{14}}\circ f_{1,u_{14}}(X) = f_{2,u_{14}}(1) = 0$ irrespective of the value of $X$. 
Each row in the table gives a unique combination of the  configuration of the variables involved, $X, Y_1, Y_2$. 
As each endogenous variable is binary, there are $2^3=8$ rows in the table. 
For a given row with configuration $(x, y_1, y_2)$, the value in cell $j$ of that row is given by 
$I\left(f_{1, u_{j}}(x)=y_1\right)\cdot I\left(f_{2, u_{j}}\circ f_{1, u_{j}}(x)=y_2\right)$. 
Seven of the eight equations are independent. 
Thus, extreme solutions $P(U)$ in $\setFont{K}(U)$ will have at most $7$ (out of 16) non-zero $P(u_i)$-values. 
By the definition of credal sets for partially specified SCMs, possible distributions $P(U)$ for observational $\tP(Y_1,Y_2|X)$ are thus given by  \cite{zaffalon2024efficient}
\begin{equation}\label{eq:cssemimarkoviandef}
\begin{aligned}    
\setFont{K}(U):= \{P(U) &:
    \sum_{u_j\in\Omega_U} P(u_j) \cdot I\left(f_{1, u_{j}}(x)=y_1\right)\cdot I\left(f_{2, u_{j}}\circ f_{1, u_{j}}(x)=y_2\right)
     \\
    &=  
    \sum_{u\in\Omega_U}[P(u)\cdot     P(Y_1, Y_2 | X, u)  ] 
    = \tP(Y_1, Y_2 | X)\}.
\end{aligned}
\end{equation}    
Table \ref{tab:semimarkovianobsmatrix} now corresponds to the credal set $\setFont{K}(U)$, in parallel to how 
Table \ref{tab:U1} defined $\setFont{K}(U_1)$ for the Markovian $U_1$.

From the representations of $u_j \in \Omega_U$ in Table \ref{tab:semimarkovianobsmatrix}, it is clear that certain 
$(f_{1,u_j}, f_{2,u_j}\circ f_{1,u_j})$ are indistinguishable from observational data. 
Consider for instance the two columns $u_{12}$ and $u_{14}$, that both dictate $(f_{1,u}(X), f_{2,u}\circ f_{1,u}(X)) = (1,0)$.
For variables $P(u_{12})$ and $P(u_{14})$ of the linear system, it follows that one is always dependent of the other, regardless of how many other variables of the system are unknown. 
Thus, in order for $P(U)$ to be an extreme solution of $\setFont{K}(U)$, at least one of $P(u_{12})$ and $P(u_{14})$ must be zero. 
By investigating Table \ref{tab:semimarkovianobsmatrix} we find that the extreme $P(U)$ for $\setFont{K}(U)$ will be solutions where at most $7$ of the $P(u_i)$ are non-zero, with at most one probability in each set of indistinguishable configurations, namely $\{P(u_0), P(u_1)\}$, $\{P(u_2), P(u_3)\}$, $\{P(u_{12}), P(u_{14})\}$, $\{P(u_{13}), P(u_{15})\}$, being non-zero.

\subsection{Experimental data}
\label{sec:semimarkovianexp}

So far, the data $\mcD$ has been assumed to be observational, and encoded through $\tP(X,Y_1,Y_2)$. 
If instead experimental data is available, then the model's credal set may be further restricted, and observationally equivalent $U$ values become distinguishable. 
Let us again consider the two states $u_{12}$ and $u_{14}$. 
They are indistinguishable from observational data because while they differ in their definition of $f_2(y_1=0)$, this distinction will not be observable since $f_{2,u_{12}}(1)=f_{2,u_{14}}(1)$ and $Y_1=1$ holds both when $U=u_{12}$ and when $U=u_{14}$.
The only way to infer their difference is by assuming that we have access to \textit{experimental} data; notice how responses on $Y_2$ on the intervention $\dop(Y_1=0)$ differ. 
This is showcased in Table \ref{tab:Udomaindef}. The matrices are defined using 
$I\left(f_{1, u_{j}}(x)=y_1\right)$ (upper part) and $I\left(f_{2, u_{j}}(y_1)=y_2\right)$ (lower part).
For simplicity of presentation, we consider the observations of $Y_1$ given its endogenous parent as an interventional observation and write $P(Y_1 | \dop(x))$, even if this probability under the considered model is identical to the observational distribution $\tP(Y_1 | x)$; cf.\ Table \ref{tab:U1}.

\begin{center}
\begin{sidewaystable}[htp]
\scriptsize
\centering
\begin{tabular}{ | p{15pt} |p{15pt}||p{9pt}|p{9pt}|p{9pt}|p{9pt}|p{9pt}|p{9pt}|p{9pt}|p{9pt}|p{9pt}|p{9pt}|p{11pt}|p{11pt}|p{11pt}|p{11pt}|p{11pt}|p{11pt}||c| } 
 \hline
  \multicolumn{2}{|c||}{$ $} &\multicolumn{16}{c||}{$U$} & \multirow{4}{4em}{$ $}\\ 
 \cline{3-18}
   \multicolumn{2}{|c||}{$ $} &  $u_0$ & $u_1$ & $u_2$ & $u_3$ & $u_4$ & $u_5$ & $u_6$ & $u_7$ & $u_8$ & $u_9$ & $u_{10}$ & $u_{11}$ & $u_{12}$ & $u_{13}$ & $u_{14}$ & $u_{15}$ & \\
\cline{1-18}
  \multicolumn{2}{|c||}{$f_{1,U}(X)= $} & 0 &0  & 0 & 0 & $X$ & $X$ & $X$ & $X$ & $\overline{X}$ & $\overline{X}$ &  $\overline{X}$& $\overline{X}$ & 1 & 1 & 1 & 1 &  $ $\phantom{$X^{X^{X^X}}$} \\

   \cline{1-18}
    \multicolumn{2}{|c||}{$f_{2,U}(Y_1)=$} & 0 & $Y_1$ & $\overline{Y_1}$ & 1 & 0 & $Y_1$ & $\overline{Y_1}$ & 1  & 0 & $Y_1$ & $\overline{Y_1}$ & 1 &0 & $Y_1$ & $\overline{Y_1}$ & 1  & $ $\phantom{$X^{X^{X^X}}$}  \\

 \hline
 \hline

 \multirow{2}{7pt}{$X$} & \multirow{2}{7pt}{$Y_1$}&  \multicolumn{16}{c||}{$ $} & \multirow{2}{4em}{$\tP(Y_1 | \dop(x))$}\\ 
  & &  \multicolumn{16}{c||}{$ $} & \\

 \hline
 \hline
 
  0 & 0 &  1 & 1 & 1 & 1 & 1 & 1 & 1 & 1 & 0 & 0 & 0 & 0 & 0 & 0 & 0 &  0 & \tiny $\tP(Y_1=0| \dop(X=0))$\\
  0 & 1 &  0 & 0 & 0 & 0 & 0 & 0 & 0 & 0 & 1 & 1 & 1 & 1 & 1 & 1 & 1 &  1  & \tiny $\tP(Y_1=1| \dop(X=0))$\\
  \hline
  1 & 0 &  1 & 1 & 1 & 1 & 0 & 0 & 0 & 0 & 1 & 1 & 1 & 1 & 0 & 0 & 0 &  0  & \tiny $\tP(Y_1=0| \dop(X=1))$\\
  1 & 1 &  0 & 0 & 0 & 0 & 1 & 1 & 1 & 1 & 0 & 0 & 0 & 0 & 1 & 1 & 1 &  1  & \tiny $\tP(Y_1=1| \dop(X=1))$\\
 \hline
\hline
  \multirow{2}{7pt}{$Y_1$} & \multirow{2}{7pt}{$Y_2$} &  \multicolumn{16}{c||}{$ $} & \multirow{2}{4em}{$\tP(Y_2 | \dop(y_1))$}\\ 
 & &  \multicolumn{16}{c||}{$ $} & \\
  
 \hline
  \hline
  0 & 0 & 1 & 1 & 0 & 0 & 1 & 1 & 0 & 0 & 1 & 1 & 0 & 0 & 1 & 1 & 0 & 0 & \tiny $\tP(Y_2=0| \dop(Y_1=0))$\\
  0 & 1 & 0 & 0 & 1 & 1 & 0 & 0 & 1 & 1 & 0 & 0 & 1 & 1 & 0 & 0 & 1 & 1 & \tiny $\tP(Y_2=1| \dop(Y_1=0))$\\
  \hline
  1 & 0 & 1 & 0 & 1 & 0 & 1 & 0 & 1 & 0 & 1 & 0 & 1 & 0 & 1 & 0 & 1 & 0  & \tiny $\tP(Y_2=0| \dop(Y_1=1))$\\
  1 & 1 & 0 & 1 & 0 & 1 & 0 & 1 & 0 & 1 & 0 & 1 & 0 & 1 & 0 & 1 & 0 & 1  & \tiny $\tP(Y_2=1| \dop(Y_1=1))$\\
 \hline
\end{tabular}
\caption{Definition of the $U$ domain, each value $u_j$ indexing a distinct combination of functions $f_{1}, f_{2}$. 
For values $(u, x, y_{1}, y_{2})$, table entries are defined by $I(f_{1,u}(x)=y_1)$ (upper part) and $I(f_{2,u}(y_1)=y_2)$ (lower part).} 
\label{tab:Udomaindef}
\end{sidewaystable}
\end{center}

Tables \ref{tab:U1} and \ref{tab:U2} were seen to correspond directly to the credal sets $\setFont{K}(U_1)$, $\setFont{K}(U_2)$ as defined for Markovian exogenous variables in partially specified SCMs. Analogously extracting a credal set based on Table \ref{tab:Udomaindef} gives
\begin{equation}\label{eq:csudomaindef}
\begin{aligned}
    \setFont{K}'(U):= \{P(U): & \sum_{u\in\Omega_U} P(u)\cdot I(f_{1,u}(x)=y_1) = 
    \tP(Y_1|\dop(X)),\\
    & \sum_{u\in\Omega_U} P(u)\cdot I(f_{2,u}(y_1)=y_2) = \tP(Y_2|\dop(Y_1)) \}  .
\end{aligned}
\end{equation}
The two credal sets $\setFont{K}(U)$ and $\setFont{K}'(U)$ (defined in Equations  \eqref{eq:cssemimarkoviandef} and \eqref{eq:csudomaindef}, respectively) are not the same. 
This invites us to combine the two approaches: The second half of credal set $\setFont{K}'(U)$ (Equation \ref{eq:csudomaindef}), namely $\sum_{u\in\Omega_U} [P(u)\cdot P(Y_2|Y_1,u)] = P(Y_2|\dop(Y_1))$, may be combined with $\setFont{K}(U)$, which gives:
\begin{equation}\label{eq:csinterventional}
\begin{aligned}
    \setFont{K}^*(U):= \{P(U): & \sum_{u\in\Omega_U} [P(u)\cdot P(Y_1, Y_2|X,u)] = \tP(Y_1, Y_2|X),\\
    & \sum_{u\in\Omega_U} [P(u)\cdot P(Y_2|Y_1,u)] = \tP(Y_2|\dop(Y_1)) \}.
\end{aligned}
\end{equation}


\begin{center}
\begin{sidewaystable}[htp]
\scriptsize
\centering
\begin{tabular}{ | p{17pt} | p{17pt} |p{17pt}||p{9pt}|p{9pt}|p{9pt}|p{9pt}|p{9pt}|p{9pt}|p{9pt}|p{9pt}|p{9pt}|p{11pt}|p{11pt}|p{11pt}|p{11pt}|p{11pt}|p{11pt}|p{11pt}||c| } 
 \hline
  \multicolumn{3}{|c||}{$ $}& \multicolumn{16}{|c||}{$U$} & \\ 
 \cline{4-19}
    \multicolumn{3}{|c||}{$ $} & $u_0$ & $u_1$ & $u_2$ & $u_3$ & $u_4$ & $u_5$ & $u_6$ & $u_7$ & $u_8$ & $u_9$ & $u_{10}$ & $u_{11}$ & $u_{12}$ & $u_{13}$ & $u_{14}$ & $u_{15}$ & \\
\cline{1-19}
 \multicolumn{3}{|c||}{$f_{1,U}(X) = $} & $0$ & $0$ & $0$ & $0$  & $X$ & $X$ & $X$ & $X$ & $\overline{X}$ & $\overline{X}$ & $\overline{X}$  & $\overline{X}$  & $1$  & $1$ & $1$ & $1$ & $ $\phantom{$X^{X^{X^X}}$}\\
 \cline{1-19}
 \multicolumn{3}{|c||}{$f_{2,U}\circ f_{1,U}(X) = $} & $0$ & $0$ & $1$ & $1$ & $0$ & $X$ & $\overline{X}$ & $1$ & $0$ & $\overline{X}$ & $X$ & $1$ & $0$ & $1$ & $0$ &  $1$&$ $\phantom{$X^{X^{X^X}}$}  \\
  \cline{1-19}
 \multicolumn{3}{|c||}{$f_{2,U}(Y_1)= $} & $0$ & $Y_1$ & $\overline{Y_1}$ & $1$ &  $0$ & $Y_1$ & $\overline{Y_1}$ & $1$&  $0$ & $Y_1$ & $\overline{Y_1}$ & $1$&  $0$ & $Y_1$ & $\overline{Y_1}$ & $1$ &$ $\phantom{$X^{X^{X^X}}$}  \\
    
 \hline
 \hline
 \multirow{2}{7pt}{$X$} & \multirow{2}{7pt}{$Y_1$}&  \multirow{2}{7pt}{$Y_2$}&\multicolumn{16}{c||}{$ $} & \multirow{2}{4em}{Probability}\\ 
  & & & \multicolumn{16}{c||}{$ $} & \\
  \hline
  \hline
  0 & 0 & 0 & 1 & 1 & 0 & 0 & 1 & 1 & 0 & 0 & 0 & 0 & 0 & 0 & 0 & 0 & 0 & 0 & \tiny $\tP(Y_1=0, Y_2=0| X=0)$\\
  0 & 0 & 1 & 0 & 0 & 1 & 1 & 0 & 0 & 1 & 1 & 0 & 0 & 0 & 0 & 0 & 0 & 0 & 0  & \tiny $\tP(Y_1=0, Y2=1| X=0)$\\
  0 & 1 & 0 & 0 & 0 & 0 & 0 & 0 & 0 & 0 & 0 & 1 & 0 & 1 & 0 & 1 & 0 & 1 & 0  & \tiny $\tP(Y_1=1, Y_2=0| X=0)$\\
  0 & 1 & 1 & 0 & 0 & 0 & 0 & 0 & 0 & 0 & 0 & 0 & 1 & 0 & 1 & 0 & 1 & 0 & 1  & \tiny $\tP(Y_1=1, Y_2=1| X=0)$\\
  \hline
  1 & 0 & 0 & 1 & 1 & 0 & 0 & 0 & 0 & 0 & 0 & 1 & 1 & 0 & 0 &  0 & 0 & 0 & 0 & \tiny $\tP(Y_1=0, Y_2=0| X=1)$\\
  1 & 0 & 1 & 0 & 0 & 1 & 1 & 0 & 0 & 0 & 0 & 0 & 0 & 1 & 1 & 0 & 0 & 0 & 0  & \tiny $\tP(Y_1=0, Y2=1| X=1)$\\
  1 & 1 & 0 & 0 & 0 & 0 & 0 & 1 & 0 &1 & 0 & 0 & 0 & 0 & 0 & 1 & 0 & 1 & 0  & \tiny $\tP(Y_1=1, Y_2=0| X=1)$\\
  1 & 1 & 1 & 0 & 0 & 0 & 0 & 0 & 1 & 0 & 1 & 0 & 0 & 0 & 0 & 0 & 1 & 0 & 1  & \tiny $\tP(Y_1=1, Y_2=1| X=1)$\\
 \hline
   - & 0 & 0 & 1 & 1 & 0 & 0 & 1 & 1 & 0 & 0 & 1 & 1 & 0 & 0 &  1 & 1 & 0 & 0 & \tiny $\tP(Y_2=0| \dop(Y_1=0))$\\
   - & 0 & 1 & 0 & 0 & 1 & 1 & 0 & 0 & 1 & 1 & 0 & 0 & 1 & 1 & 0 & 0 & 1 & 1  & \tiny $\tP(Y2=1| \dop(Y_1=1))$\\
   - & 1 & 0 & 1 & 0 & 1 & 0 & 1 & 0 &1 & 0 & 1 & 0 & 1 & 0 & 1 & 0 & 1 & 0  & \tiny $\tP(Y_2=0| \dop(Y_1=0))$\\
   - & 1 & 1 & 0 & 1 & 0 & 1 & 0 & 1 & 0 & 1 & 0 & 1 & 0 & 1& 0 & 1 & 0 & 1  & \tiny $\tP(Y_2=1| \dop(Y_1=1))$\\
  \hline

\end{tabular}
\caption{The definition of the $U$-domain with observational data (Table \ref{tab:semimarkovianobsmatrix}) is extended with experimental data.}
\label{tab:semimarkovianintmatrix}
\end{sidewaystable}
\end{center}

Now each $u_i$ is contributing to a distinct set of equations. As more independent equations are included while $\Omega_U$ remains the same, $\setFont{K}^*(U) \subseteq \setFont{K}(U)$. 

With access to complete experimental data, all interventional queries are identifiable. Counterfactual queries may be bounded, possibly at better precision given the additional information provided about $\Omega_U$. 






    

\section{Markovian approximations}
\label{sec:markovianapprox}

Query calculations in semi-Markovian models may be approximated with Markovian models. Two approaches are discussed here, where minimal adaptations to the model are made such that Markovian DCCC may be applied. The first approach relaxes the conditional independence such that $X \centernot\perp\!\!\!\perp Y_2 | Y_1, U$ and is detailed in Section \ref{sec:endogenousmerge}. The second approach enforces stronger conditional independence $X \perp\!\!\!\perp Y_2 | Y_1 $, and is presented in Section~\ref{sec:exogenoussplit}.

\subsection{Endogenous merge}
\label{sec:endogenousmerge}

\begin{figure}[ht]
\centering
\begin{tikzpicture}[thick, main/.style = {draw, circle}]

\node[dot,label=below:{$X$}] (X)  at (-1.5,0) {};
\node[dot,label=below:{$Y$}] (Y)  at (0,0) {};		
\node[dot2,label=above :{$U_0$}] (V)  at (-1.5,1.5) {};
\node[dot2,label=above :{$U^*$}] (U)  at (0,1.5) {};
\draw[a2] (X) -- (Y);
\draw[a] (V) -- (X);
\draw[a] (U) -- (Y);




\end{tikzpicture}

\caption{Markovian approximation of Figure \ref{fig:modelexample}, replacing variables $Y_1, Y_2$ that share exogenous parent $U$ by new variable $Y = (Y_1, Y_2)$ with domain $\Omega_{Y} = \{00, 01, 10, 11\}$ and new exogenous parent $U^*$.}
\label{fig:modelexample_markov}
\end{figure}
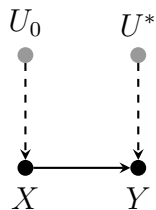

The first Markovian approximation considered is based on merging the endogenous variables that share a confounding exogenous parent, such that these combined are considered the single endogenous child of the exogenous variable, in line with the Markovian assumption. 
For the example model in Figure \ref{fig:modelexample}, this corresponds to replacing variables $Y_1, Y_2$ that share the exogenous parent $U$ by the new variable $Y = (Y_1, Y_2)$, and assigning $Y$ the exogenous parent $U^*$. 
This renders the model Markovian, see Figure \ref{fig:modelexample_markov}. The Markovian SCM is further defined by letting
$\Omega_{Y} = \{00, 01, 10, 11\}$, and replacing the two functions $Y_1 = f_1(X, U)$ and $Y_2 = f_2(Y_1, U)$ by $(Y_1, Y_2) = f_3(X, U^*) = f_{3,U^*}(X)$.

This model now being Markovian, the canonical specification $\Omega_{U^*}$ of $U^*$ has domain size $|\Omega_{U^*}| = |\Omega_{Y}|^{|\Omega_{X}|} = 16$, indexing possible functions $f_{3, u^*_i}$ as shown in Table \ref{tab:markovianapprox}. Note that it is not the case that Markovian $\Omega^*_{U}$ equals the original semi-Markovian canonical domain $\Omega_{U}$, as the canonical set of functions $f_{3,U^*}$ is not the same as the semi-Markovian $(f_{1,U}, f_{2,U}\circ f_{1,U})$ counterpart. This is exemplified by considering $u^*_1$ in Table \ref{tab:markovianapprox}, which indexes a function that is not allowed by the semi-Markovian definition, that is $f_{3, u^*_1}(X) = (0, X)$. Here, $Y_1$ is always at 0, while $Y_2 = X$. This allows influence from $X$ directly on $Y_2$, which is known not to be the case for the original model. As the specification for $f_{3,U^*}$ is canonical, it will however contain all $(f_{1,U}, f_{2,U}\circ f_{1,U})$ distinguishable by observation, and is thus a more general specification. We will revisit this result in Section \ref{sec:markovianapproxexperiments}. 



\begin{center}
\begin{sidewaystable}[htp]
\scriptsize
\centering
\begin{tabular}{ | p{10pt} | p{10pt} |p{10pt}||p{11pt}|p{11pt}|p{11pt}|p{11pt}|p{11pt}|p{11pt}|p{11pt}|p{11pt}|p{11pt}|p{11pt}|p{11pt}|p{11pt}|p{11pt}|p{11pt}|p{11pt}|p{11pt}||c| } 
 \hline
  \multicolumn{3}{|c||}{$ $} & \multicolumn{16}{|c||}{$U^*$} & \\ 
 \cline{4-19}
   \multicolumn{3}{|c||}{$ $} & $u^*_0$ & $u^*_1$ & $u^*_2$ & $u^*_3$ & $u^*_4$ & $u^*_5$ & $u^*_6$ & $u^*_7$ & $u^*_8$ & $u^*_9$ & $u^*_{10}$ & $u^*_{11}$ & $u^*_{12}$ & $u^*_{13}$ & $u^*_{14}$ & $u^*_{15}$ & \\
 \cline{1-19}
     \multicolumn{3}{|c||}{$f_{3,U^*}(X) = $} & $(0,$ & $(0,$ & $(X,$ & $(X,$  & $(0,$ & $(0,$ & $(X,$ & $(X,$ & $(\overline{X},$ & $(\overline{X},$ & $(1,$ & $(1,$ & $(\overline{X},$  & $(\overline{X},$ & $(1,$ & $(1,$ & \\
 \multicolumn{3}{|c||}{$ $} & $\;\;0)$ & $\;X)$ & $\;\;0)$ & $\;X)$ & $\;\overline{X})$ & $\;\;1)$ & $\;\overline{X})$ & $\;\;1)$ & $\;\;0)$ & $\;X)$ & $\;\;0)$ & $\;X)$ & $\;\overline{X})$ & $\;\;1)$ & $\;\overline{X})$ &  $\;\;1)$&  \\
 
 \hline
 \hline
 \multirow{2}{7pt}{$X$} & \multirow{2}{7pt}{$Y_1$}&  \multirow{2}{7pt}{$Y_2$}&\multicolumn{16}{c||}{$ $} & \multirow{2}{4em}{$\tP(Y_1, Y_2 | X)$}\\ 
  & & & \multicolumn{16}{c||}{$ $} & \\
  \hline
  \hline

  0 & 0 & 0 & 1 & 1 & 1 & 1 & 0 & 0 & 0 & 0 & 0 & 0 & 0 & 0 & 0 & 0 & 0 & 0 & \tiny $P(Y_1=0, Y_2=0| X=0)$\\
  0 & 0 & 1 & 0 & 0 & 0 & 0 & 1 & 1 & 1 & 1 & 0 & 0 & 0 & 0 & 0 & 0 & 0 & 0  & \tiny $P(Y_1=0, Y2=1| X=0)$\\
  0 & 1 & 0 & 0 & 0 & 0 & 0 & 0 & 0 & 0 & 0 & 1 & 1 & 1 & 1 & 0 & 0 & 0 & 0  & \tiny $P(Y_1=1, Y_2=0| X=0)$\\
  0 & 1 & 1 & 0 & 0 & 0 & 0 & 0 & 0 & 0 & 0 & 0 & 0 & 0 & 0 & 1 & 1 & 1 & 1  & \tiny $P(Y_1=1, Y_2=1| X=0)$\\
  \hline
  1 & 0 & 0 & 1 & 0 & 0 & 0 & 1 & 0 & 0 & 0 & 1 & 0 & 0 & 0 & 1 & 0 & 0 & 0 & \tiny $P(Y_1=0, Y_2=0| X=1)$\\
  1 & 0 & 1 & 0 & 1 & 0 & 0 & 0 & 1 & 0 & 0 & 0 & 1 & 0 & 0 & 0 & 1 & 0 & 0  & \tiny $P(Y_1=0, Y2=1| X=1)$\\
  1 & 1 & 0 & 0 & 0 & 1 & 0 & 0 & 0 & 1 & 0 & 0 & 0 & 1 & 0 & 0 & 0 & 1 & 0  & \tiny $P(Y_1=1, Y_2=0| X=1)$\\
  1 & 1 & 1 & 0 & 0 & 0 & 1 & 0 & 0 & 0 & 1 & 0 & 0 & 0 & 1 & 0 & 0 & 0 & 1  & \tiny $P(Y_1=1, Y_2=1| X=1)$\\
 \hline

 \hline
\end{tabular}
\caption{The canonical specification of the $U^*$-domain when the endogenous child $Y$ of $U^*$ has 4 states and $Y$'s endogenous parent $X$ is binary.}
\label{tab:markovianapprox}
\end{sidewaystable}
\end{center}



Now, Markovian DCCC is directly applicable solving the equations in Table \ref{tab:markovianapprox} for observed $\tP(Y_1, Y_2|X)$. This corresponds to finding the extreme points of the Markovian equivalent $\setFont{K}(U^*)$ of the credal set $\setFont{K}(U)$
\begin{equation}\label{eq:csmarkovianapprox}
    \setFont{K}(U^*):= \left\{P(U^*):\sum_{u\in\Omega_{U^*}}[P(u)\cdot P(Y | X, u) ] = P(Y_1, Y_2 | X)\right\} .
\end{equation}

The Markovian approximation being a more general model specification, bounding queries will ensure query intervals under the approximate model always contain the true query interval under the semi-Markovian model. The approximate model however now only permits interventions on endogenous variables $X$ and $Y$, the latter corresponding to joint interventions on $Y_1, Y_2$. Specifically, the ability to monitor the effect of $Y_1$ on $Y_2$ is lost. Essentially, the characteristics of Markovian partially specified SCMs are in effect: Interventional queries in the new set of variables $X, Y$ are identifiable, whereas counterfactual queries conditioned on interventions in $X, Y$ may at best be expressed by interval bounds. 


\subsection{Exogenous split}
\label{sec:exogenoussplit}

A second alternative for Markovian approximation is to assume the model structure shown in Figure \ref{fig:examplemodel_markovian} as a simplification, even if the data suggests $X \centernot {\perp\!\!\!\perp} Y_2 | Y_1$. The query intervals are calculated as if the exogenous influence on $Y_1$ is independent of that on $Y_2$. This approximation approach is named exogenous split, referencing the model adaptation where a confounding exogenous variable $U$ of a semi-Markovian model is replaced by a pair of independent exogenous variables $U_1, U_2$. As this approximate model introduces new conditional dependence assumptions, approximate query intervals are no longer guaranteed to contain the true interval.


\section{Experiments}\label{sec:experiments}

Several experiments with DCCC for semi-Markovian models are presented here. All experiments consider the semi-Markovian model discussed in Section \ref{sec:semimarkovian}, with topology as seen in Figure \ref{fig:modelexample}, or its Markovian approximations (Section \ref{sec:markovianapprox}). All experiments consider canonical models.

Section \ref{sec:semimarkoviandataexperiments} details experiments comparing query intervals for semi-Marko\-vian models with various data sources. Section \ref{sec:markovianapproxexperiments} presents experiments comparing intervals retrieved with semi-Markovian DCCC with intervals retrieved with Markovian approximation methods. Section \ref{sec:literaturemethodcomparison} presents a comparison of semi-Markovian DCCC with other methods for interval computation from the literature. 

\subsection{Semi-Markovian query bounding with observational and experimental data}
\label{sec:semimarkoviandataexperiments}

Experiments are presented comparing the impact on query intervals of various data sources in semi-Markovian models.

For the model in question (Figure \ref{fig:modelexample}), a model specific implementation of DCCC is considered, where the linear system to be solved is made explicit. For observational data $\tP(Y_1, Y_2, X)$, this system corresponds to Table \ref{tab:semimarkovianobsmatrix} and credal set $\setFont{K}(U)$ (Equation \ref{eq:cssemimarkoviandef}), which has 7 independent equations. Exhaustive search is performed considering all $\binom{16}{7}$ possible subsets of $\Omega_U$ as possible extreme points, retaining all solutions corresponding to probability distributions. In experiment results, this approach is named S-O. 

If experimental data $\tP(Y_2 | \dop(Y_1))$ is also available, the linear system is the one shown in Table \ref{tab:semimarkovianintmatrix} and credal set $\setFont{K}^*(U)$ (Equation \ref{eq:csinterventional}), which now has 9 independent equations. In experiment results, this approach is named S-OE.

Finally, while less likely to be the case in practice, the case where only experimental data is available is considered for completeness. The linear system corresponds to Table \ref{tab:Udomaindef} and credal set $\setFont{K}'(U)$ (Equation \ref{eq:csudomaindef}), and has 5 independent equations. Note that in this case, the model specific implementation simplifies to a slight adaptation of the Markovian DCCC. The structure of the linear system with unknowns $\{P(u_i)\}_{i=0}^{15}$ is in this case identical to that of a Markovian model $X' \rightarrow Y' \leftarrow U'$ with $|\Omega_{Y'}| = 2$ and $|\Omega_{X'}| = 4$, and may thus be solved using the Markovian DCCC with distributions $\tP(Y_2|\dop(Y_1))$, $\tP(Y_1|\dop(X))$ in place of $P(Y'|X')$. Note that with only experimental data, no information is provided about how $X$ and $Y_2$ correlate. In experiment results, this approach is named S-E. 

Table \ref{tab:semimarkovianvariants} summarises the approaches compared in the experiments discussed in this section.

\begin{table}[ht]
\scriptsize
\centering
\begin{tabular}{  p{3.2cm} | p{1.0cm} | p{2.5cm } | p{2cm} | p{2.5cm} }
 Approach & Credal set & Graph & Dataset $\mcD$ & Solution search \\
 \hline
  
  Semi-Markovian model, Observational data (S-O) (Section \ref{sec:semimarkovianobs})& $\setFont{K}(U)$ & \begin{tikzpicture}[scale=0.5]
			\node[dot,label=below:{$X$}] (X)  at (-1.5,0) {};
			\node[dot,label=below:{$Y_1$}] (Y1)  at (0,0) {};		
            \node[dot,label=below:{$Y_2$}] (Y2)  at (1.5,0) {};
			\node[dot2,label=above :{$U_0$}] (V)  at (-1.5,1.5) {};
			\node[dot2,label=above :{$U$}] (U)  at (0.75,1.5) {};
			\draw[a2] (X) -- (Y1);
            \draw[a2] (Y1) -- (Y2);
			\draw[a] (V) -- (X);
			\draw[a] (U) -- (Y1);
            \draw[a] (U) -- (Y2);     
		\end{tikzpicture}   & $\tP(Y_1, Y_2,X)$ & Semi-Markovian exhaustive DCCC for exogenous $U$ - Model specific implementation \\
  \hline
  
  Semi-Markovian model, Observational and Experimental data (S-OE) (Section \ref{sec:semimarkovianexp})& $\setFont{K}^*(U)$ & \begin{tikzpicture}[scale=0.5]
			\node[dot,label=below:{$X$}] (X)  at (-1.5,0) {};
			\node[dot,label=below:{$Y_1$}] (Y1)  at (0,0) {};		
            \node[dot,label=below:{$Y_2$}] (Y2)  at (1.5,0) {};
			\node[dot2,label=above :{$U_0$}] (V)  at (-1.5,1.5) {};
			\node[dot2,label=above :{$U$}] (U)  at (0.75,1.5) {};
			\draw[a2] (X) -- (Y1);
            \draw[a2] (Y1) -- (Y2);
			\draw[a] (V) -- (X);
			\draw[a] (U) -- (Y1);
            \draw[a] (U) -- (Y2);     
		\end{tikzpicture}  & $\tP(Y_1, Y_2,X),$ $\tP(Y_2|\dop(Y_1))$ & Semi-Markovian exhaustive DCCC for exogenous $U$ - Model specific implementation\\

  \hline
  
  Semi-Markovian model, Experimental data (S-E) (Section \ref{sec:semimarkovianexp})& $\setFont{K}'(U)$ & \begin{tikzpicture}[scale=0.5]
			\node[dot,label=below:{$X$}] (X)  at (-1.5,0) {};
			\node[dot,label=below:{$Y_1$}] (Y1)  at (0,0) {};		
            \node[dot,label=below:{$Y_2$}] (Y2)  at (1.5,0) {};
			\node[dot2,label=above :{$U_0$}] (V)  at (-1.5,1.5) {};
			\node[dot2,label=above :{$U$}] (U)  at (0.75,1.5) {};
			\draw[a2] (X) -- (Y1);
            \draw[a2] (Y1) -- (Y2);
			\draw[a] (V) -- (X);
			\draw[a] (U) -- (Y1);
            \draw[a] (U) -- (Y2);     
		\end{tikzpicture} & $\tP(X)$, $ \tP(Y_1|\dop(X)),$ $\tP(Y_2|\dop(Y_1))$& Markovian exhaustive DCCC adapted for experimental distributions  
        \\

  
  
  \hline
  
 \end{tabular}
 \caption{Semi-Markovian approaches to computing query intervals.}
 \label{tab:semimarkovianvariants}
 \end{table}

By definition, the credal sets of the semi-Markovian approaches relate in the following way: $\setFont{K}^*(U) \subseteq \setFont{K}(U)$ and $\setFont{K}^*(U) \subseteq \setFont{K}'(U)$. From this, it follows that for exact intervals computed for any query:

$$ \text{S-OE interval} \subseteq \text{S-O interval},$$
$$ \text{S-OE interval} \subseteq \text{S-E interval} .$$

For the experiments, we will consider queries related to the so-called \textit{probability of necessity and sufficiency} which is defined, for two given variables $X$ and $Y$, as~\cite{pearl2009}
\begin{equation}
\label{eq:pns}
    \mathrm{PNS}(X,Y) = P(Y_{x}=y , Y_{x'}=y') ,
\end{equation}
and it measures how $Y$ reacts to $X$, hence expressing  to what extent $X=x$ is necessary and sufficient for $Y=y$. The notation $Y_{x}$ denotes the variable $Y$ in the hypothetical scenario where the variable $X$ is intervened to take the value $x$, given that in reality, events $x'$ and $y'$ occurred.
Exact intervals are computed for the three queries PNS$(X,Y_1)$, PNS$(X,Y_2)$ and PNS$(Y_1, Y_2)$ for 500 randomly generated distributions $\tP(Y_1, Y_2, X)$ and $\tP(Y_2|Y_1)$, with exhaustive DCCC for each of the three semi-Markovian approaches. Table \ref{tab:smavglen} details the average length of the query intervals retrieved. S-OE intervals are on average shown to be smaller than both S-O and S-E intervals for all queries. The experimental data approach S-E is the least informative for both queries involving interventions on $X$. This is in particular the case for PNS$(X, Y_2)$, which is reasonable as no information was provided on the co-occurrence of variables $X,Y_2$. For the query PNS$(Y_1, Y_2)$ however, S-E is seen to perform better than S-O on average, with interventions on $Y_1$ being a shortcoming in the observational setting.  

\begin{table}
\centering
\begin{tabular}{ |c|c|c|c| } 
 \hline
 $ $ & S-OE & S-O & S-E \\ 
 \hline
 PNS$(X,Y_1)$ & 0.361 & 0.388 & 0.417 \\
 \hline
 PNS$(X,Y_2)$ & 0.350 & 0.383 & 0.848 \\
 \hline
 PNS$(Y_1,Y_2)$ & 0.361 & 0.470 & 0.418\\

\hline
\end{tabular}
\caption{Average length of the intervals retrieved for the selected queries, for 500 randomly generated distributions $\tP(Y_1, Y_2, X)$ and $\tP(Y_2|\dop(Y_1))$. }
\label{tab:smavglen}
\end{table}

The comparison of intervals is further explored in Tables \ref{tab:smintervalcomparison} and \ref{tab:smintervalcomparison2}. Here, for each pair of approaches S-OE, S-O and S-E, the number of equal intervals are highlighted, illustrating that even for query PNS$(X,Y_1)$, involving variables $Y_1, X$ for which $P(Y_1|X) = P(Y_1|\dop(X))$, S-OE returns an interval strictly contained in that of S-O for 65.8 \% of the models, and 89.0 \% of the models when compared to S-E. For the other queries, these percentages are even higher. Note that Table \ref{tab:smintervalcomparison2} reveals that while S-O intervals $\subseteq$ S-E intervals for all models for both queries PNS$(X,Y_1)$ and PNS$(X,Y_2)$, it is not the case that S-E intervals $\subseteq$ S-O intervals for all models for the query PNS$(Y_1, Y_2)$ with intervention on $Y_1$, and thus it is model dependent whether observational or experimental data provides the most information about this particular query. 

\begin{table}
\centering
\scalebox{0.73}{
\begin{tabular}{ |c|c|c|c||c|c|c| } 
 \hline
 & S-OE$=$S-O & S-OE $\subset$ S-O & S-OE $\not\subseteq$ S-O  & S-OE $=$ S-E & S-OE $\subset$ S-E & S-OE $\not\subseteq$ S-E \\ 
 \hline
 PNS$(X,Y_1)$ & 34.2 \% & 65.8 \%& 0.0 \%& 11.0 \% & 89.0 \%& 0.0 \% \\
 \hline
 PNS$(X,Y_2)$ & 19.4 \% & 80.6\% & 0.0 \%& 0.0 \%& 100.0\% & 0.0 \% \\
 \hline
 PNS$(Y_1,Y_2)$ & 5.2 \% & 94.8 \% & 0.0 \%& 8.8 \%& 91.2 \%& 0.0 \% \\
 \hline
\end{tabular}
}
\caption{For each pair of methods S-OE vs S-O and S-OE vs S-E, the table details the percentage of the intervals retrieved that are equal, and the percentage of S-OE intervals strictly contained in its S-O and S-E counterparts.}
\label{tab:smintervalcomparison}
\end{table}

\begin{table}
\centering
\begin{tabular}{ |c|c|c|c|c| } 
 \hline
 & S-O $=$ S-E & S-O $\subset$ S-E & S-E $\subset$ S-O & S-O $\not\underset{\supset}{\overset{\subset}{=}}$ S-E \\ 
 \hline
 PNS$(X,Y_1)$ & 34.2 \%& 65.8 \%& 0.0 \%& 0.0 \%\\
 \hline
 PNS$(X,Y_2)$ & 0.0 \%& 100.0 \%& 0.0 \%& 0.0 \%\\
 \hline
 PNS$(Y_1,Y_2)$ & 0.0 \%& 20.4 \%& 57.2 \%& 22.4 \%\\
 \hline
\end{tabular}
\caption{Comparing the S-O and S-E approaches, the table details the percentage of intervals that are equal, the percentage for which S-E intervals $\subset$ S-O intervals, or S-O intervals $\subset$ S-E intervals. Notation $\not\underset{\supset}{\overset{\subset}{=}}$ in the rightmost column header is used to denote that neither interval is a subset of the other.}
\label{tab:smintervalcomparison2}
\end{table}

\subsection{Markovian approximations}
\label{sec:markovianapproxexperiments}

This section compares the previously introduced semi-Markovian DCCC approach with observational data (S-O), to two Markovian approximation approaches. These are the Markovian approximation with endogenous merge (MM-O), discussed in Section \ref{sec:endogenousmerge}, and the Markovian approximation with exogenous split (MS-O), discussed in Section \ref{sec:exogenoussplit}. DCCC for MM-O finds extreme points of the credal set $\setFont{K}(U^*)$ (Equation \ref{eq:csmarkovianapprox}), while DCCC for MS-O finds extreme points for both credal sets $\setFont{K}_1(U_1)$ (Equation \ref{eq:csetU1}) and $\setFont{K}_2(U_2)$ (Equation \ref{eq:csetU2}). Table \ref{tab:approxexperimentoverview} summarises the approaches compared in the experiments discussed in this section.

\begin{table}[ht]
\scriptsize
\centering
\begin{tabular}[t]{ p{3.2cm} |  p{1.0cm} | p{2.5cm } | p{2cm} | p{2.5cm} }
 Approach & Credal set & Graph & Dataset $\mcD$ & Solution search\\
 \hline
 Semi-Markovian model, Observational data (S-O) (Section \ref{sec:semimarkovianobs} )&  
 $\setFont{K}(U)$ &
\begin{tikzpicture}[scale=0.5]
    \node[dot,label=below:{$X$}] (X)  at (-1.5,0) {};
    \node[dot,label=below:{$Y_1$}] (Y1)  at (0,0) {};		
    \node[dot,label=below:{$Y_2$}] (Y2)  at (1.5,0) {};
    \node[dot2,label=above :{$U_0$}] (V)  at (-1.5,1.5) {};
    \node[dot2,label=above :{$U$}] (U)  at (0.75,1.5) {};
    \draw[a2] (X) -- (Y1);
    \draw[a2] (Y1) -- (Y2);
    \draw[a] (V) -- (X);
    \draw[a] (U) -- (Y1);
    \draw[a] (U) -- (Y2);
\end{tikzpicture}
 & 
$\tP(Y_1, Y_2, X)$  &  
Semi-Markovian exhaustive DCCC for exogenous $U$ - Model specific implementation\\
\hline
  
  Markovian Approximation w/ endogenous Merge, Observational data (MM-O) (Section \ref{sec:endogenousmerge}) & $\setFont{K}(U^*)$ & \begin{tikzpicture}[thick, main/.style = {draw, circle}, scale=0.5]

\node[dot,label=below:{$X$}] (X)  at (-1.5,0) {};
\node[dot,label=below:{$Y$}] (Y)  at (0,0) {};		
\node[dot2,label=above :{$U_0$}] (V)  at (-1.5,1.5) {};
\node[dot2,label=above :{$U^*$}] (U)  at (0,1.5) {};
\draw[a2] (X) -- (Y);
\draw[a] (V) -- (X);
\draw[a] (U) -- (Y);
\end{tikzpicture}  & $\tP(Y_1, Y_2, X)$  & Markovian exhaustive DCCC for exogenous $U^*$ - General approach \\

  \hline
  
  Markovian Approximation w/ exogenous Split - Observational data (MS-O) (Section \ref{sec:exogenoussplit}) & $\setFont{K}_1(U_1)$, $\setFont{K}_2(U_2)$ & 		\begin{tikzpicture}[scale=0.5]
			\node[dot,label=below:{$X$}] (X)  at (-1.5,0) {};
			\node[dot,label=below:{$Y_1$}] (Y1)  at (0,0) {};		
            \node[dot,label=below:{$Y_2$}] (Y2)  at (1.5,0) {};
			\node[dot2,label=above :{$U_0$}] (V)  at (-1.5,1.5) {};
			\node[dot2,label=above :{$U_1$}] (U1)  at (0,1.5) {};

            \node[dot2,label=above :{$U_2$}] (U2)  at (1.5,1.5) {};
			\draw[a2] (X) -- (Y1);
            \draw[a2] (Y1) -- (Y2);
			\draw[a] (V) -- (X);
			\draw[a] (U1) -- (Y1);
            \draw[a] (U2) -- (Y2);

		\end{tikzpicture} & $\tP(Y_1, Y_2, X)$ & Markovian exhaustive DCCC for exogenous $U_1$, $U_2$ - General approach \\ 
  
  \hline
  
 \end{tabular}
 \caption{The table details approaches to query interval computation that all consider observational data, but vary in model assumptions.}
 \label{tab:approxexperimentoverview}
 \end{table}

As discussed in Section \ref{sec:endogenousmerge}, query intervals found for $\setFont{K}(U)$ are always contained in query intervals for $\setFont{K}(U^*)$. Thus assuming exact interval computation, it holds that

$$ \text{S-O interval} \subseteq \text{MM-O interval} .$$

Here, exact intervals are computed for the queries PNS$(X,Y_1)$, PNS$(X,Y_2)$ with all three approaches, while PNS$(Y_1, Y_2)$ is calculated for S-O and MS-O.
In the case of MM-O,  PNS$(Y_1,Y_2)$ cannot be computed. Query calculations for PNS$(X,Y_1)$, PNS$(X,Y_2)$ with MM-O are included:

\begin{align*}
    \text{PNS}(X, Y_1) = \sum_{u^* \in \Omega_{U^*}}\sum_{y_2 \in \{0,1\}} [P(u^*)&\cdot P(Y_1=0, Y_2 = y_2 | X=0, u^*)\\ &\cdot P(Y_1=1, Y_2 = y_2 | X=1, u^*)]\\ = P(u^*_2) + P(u^*_3) +P(u^*_6)& +P(u^*_7) ,
\end{align*}

\begin{align*}
    \text{PNS}(X, Y_2) = \sum_{u^* \in \Omega_{U^*}}\sum_{y_1 \in \{0,1\}}  [P(u^*)&\cdot P(Y_2=0, Y_1 = y_1 | X=0, u^*) \\
    &\cdot P(Y_2=1, Y_1 = y_1 | X=1, u^*)]\\
    = P(u^*_1) + P(u^*_3) +P(u^*_9)& +P(u^*_{11}) .
\end{align*}

Queries for the other approaches are calculated conventionally. 

For the experiments, exact query intervals are computed for 500 randomly generated distributions $\tP(Y_1, Y_2, X)$ with exhaustive DCCC for each of the three approaches.

\begin{table}
\centering
\begin{tabular}{ |c|c|c|c| } 
 \hline
 $ $ & S-O & MM-O & MS-O \\ 
 \hline
 PNS$(X,Y_1)$ & 0.388 & 0.417 & 0.417 \\
 \hline
 PNS$(X,Y_2)$ & 0.383 & 0.418 & 0.403\\
 \hline
 PNS$(Y_1,Y_2)$ & 0.470 & - & 0.395 \\

\hline
\end{tabular}
\caption{Average length of the intervals retrieved for the selected queries, for 500 randomly generated distributions $\tP(Y_1, Y_2, X)$.}
\label{tab:avglenmark}
\end{table}

Table \ref{tab:avglenmark} details the average length of the query intervals retrieved. Table \ref{tab:mintervalcomparison} presents details of the frequency that intervals from different methods coincide, is strictly contained in the other intervals, or neither. It is seen that while MM-O always produces query intervals that contain the true semi-Markovian interval, this is not the case for MS-O.

\begin{figure}[ht]
    \centering
    \includegraphics[width=0.8\linewidth]{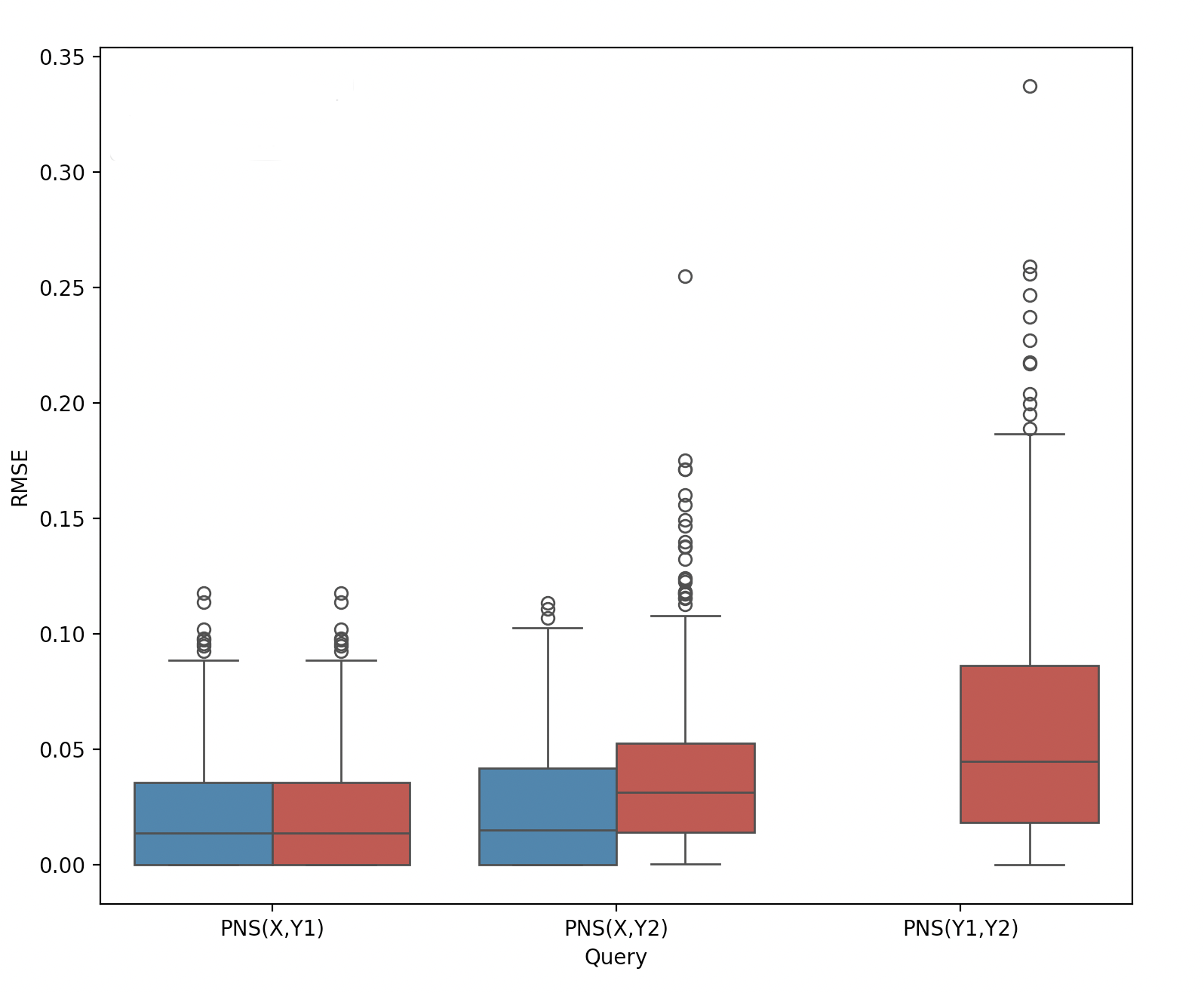}
    \caption{The figure shows boxplots of the root mean squared error (RMSE) calculated for intervals computed with S-O and MM-O in blue, and S-O and MS-O in red.}
    \label{fig:rmse}
\end{figure}

Finally, Figure \ref{fig:rmse} presents a boxplot of the root mean squared error (RMSE) of both approximations compared with the true semi-Markovian interval. While MM-O and MS-O perform comparably for PNS$(X,Y_1)$, MM-O displays lower error on the PNS$(X,Y_2)$ intervals. For PNS$(Y_1, Y_2)$, MM-O is not applicable, while MS-O displays a higher error than for the other queries. Keeping in mind that the average interval length for PNS$(Y_1, Y_2)$ with MS-O is 0.395, and this interval contains the true interval for only 14.5 \% of the models seen, this suggest low quality approximations in this case, not necessarily preferable to MM-O's inability to compute an approximation.

\begin{table}
\centering
\scalebox{.80}{
\begin{tabular}{ |c|c|c|c||c|c|c|c|} 
 \hline
 &\scriptsize  S-O$=$MM-O & \scriptsize S-O $\subset$ MM-O & \scriptsize S-O $\not\subseteq$ MM-O  & \scriptsize S-O $=$ MS-O & \scriptsize S-O $\subset$ MS-O & \scriptsize MS-O $\subset$ S-O & \scriptsize S-O $\not\underset{\supset}{\overset{\subset}{=}}$ MS-O \\ 
 \hline
 \scriptsize PNS$(X,Y_1)$ & 34.2 \% & 65.8 \% & 0.0 \% & 34.2 \% & 65.8 \% & 0.0 \% & 0.0 \% \\
 \hline
 \scriptsize PNS$(X,Y_2)$ & 35.6 \% & 64.4 \% & 0.0 \% & 0.0 \% & 23.6 \% & 14.2 \% & 62.2 \% \\
 \hline
 \scriptsize PNS$(Y_1,Y_2)$ & -  & - & - & 0.0 \% & 14.4 \% & 66.4 \% & 19.2 \% \\
 \hline
\end{tabular}
}
\caption{For each pair of methods S-O vs MM-O and S-O vs MS-O, the table details the percentage of the intervals retrieved that are equal, and the percentage of S-O intervals strictly contained in their MM-O and MS-O counterparts. The percentage of MS-O intervals contained in their S-O counterpart, as well as the percentage of cases where neither interval is a subset of the other, is also given. Notation $\not\underset{\supset}{\overset{\subset}{=}}$ denotes that neither interval is a subset of the other.}
\label{tab:mintervalcomparison}
\end{table}

\subsection{Comparison to the state of the art}
\label{sec:literaturemethodcomparison}

We compare our approach for semi-Markovian models with two methods from the literature:  
EMCC~\cite{zaffalon2024efficient}, configured with a fixed number of $100$ EM iterations; and the Gibbs sampling method in~\cite{zhang2021}, employing a burn-in period of $100$ iterations and computing the full $100\%$ credible interval. Our experiments are conducted on $100$ randomly generated SCMs, each with the topology shown in Figure~\ref{fig:modelexample}.  
All models are specified in canonical form. For each model and method, the learning performance is assessed by varying the number of generated solutions, where each solution corresponds to a fully specified SCM. We employ the same set of queries used in the previous subsection.  
The reported computation time includes both the generation of $N$ solutions and the inference required to answer one of the aforementioned queries.  
The approximation error is measured via the root mean squared error (RMSE) with respect to the exact bounds.



\begin{figure}[htbp]
    \centering
    \begin{subfigure}[b]{0.48\linewidth}
        \centering
        \includegraphics[width=\linewidth]{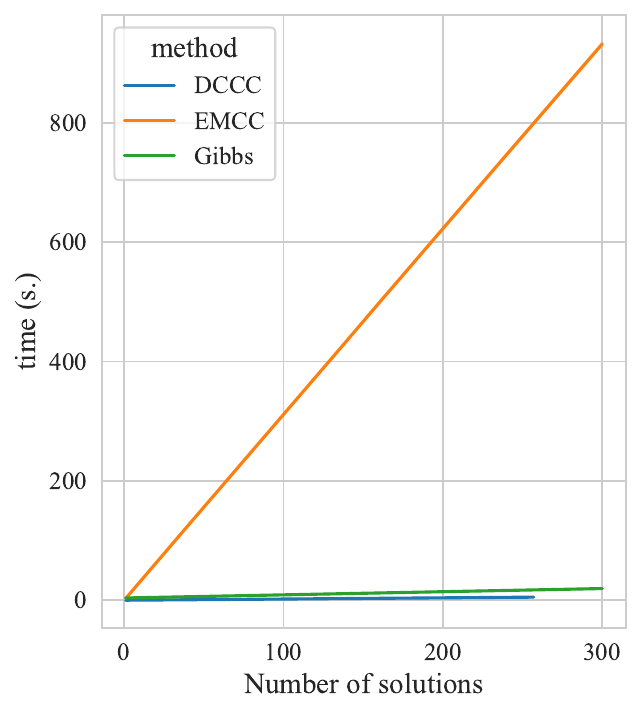}
        \label{fig:comp-time}
    \end{subfigure}
    \hfill
    \begin{subfigure}[b]{0.48\linewidth}
        \centering
        \includegraphics[width=\linewidth]{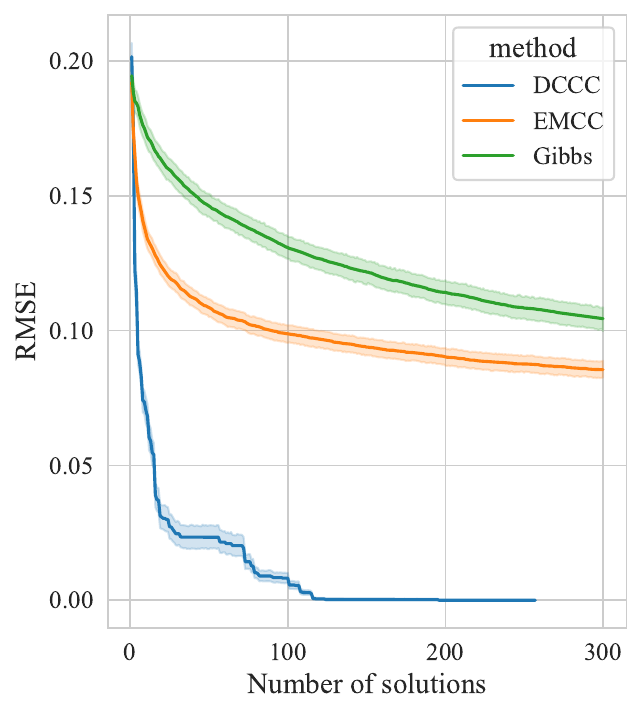}
        \label{fig:comp-error}
    \end{subfigure}
    \caption{Comparison in terms of (left) computation time and (right) error as a function of the number of generated solutions.}
    \label{fig:comparison}
\end{figure}

Figure~\ref{fig:comparison} (left) shows the average computation time in seconds (including learning and inference) vs.\ the number of generated solutions.  
All computations were carried out on a computing cluster with $1024$ cores (AMD EPYC 7542 32-Core Processor), where each experiment was executed sequentially on a single core.  
The results indicate that DCCC is the most efficient method for generating solutions, with performance comparable to Gibbs sampling.  
In contrast, EMCC is substantially less efficient.
Considering the error depicted in Figure~\ref{fig:comparison} (right), we observe that, for the same number of solutions, our method consistently yields a lower approximation error.  
Furthermore, DCCC requires slightly more than $100$ solutions to recover the exact result.  
Note that the series for this method terminates at $256$ solutions, as at this point DCCC has enumerated all possible extreme points.

\begin{figure}
    \centering
    \includegraphics[width=0.9\linewidth]{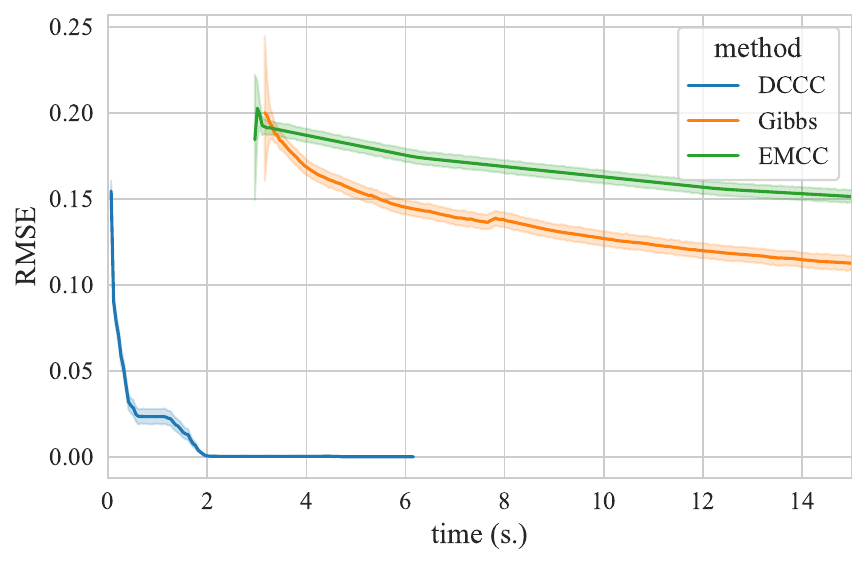}
    \caption{Error against computation time. 
    }
    \label{fig:errortime}
\end{figure}

Although EMCC and Gibbs could, in principle, achieve a comparable error given an extremely large number of iterations, such an approach would be prohibitively time-consuming.  
It is therefore instructive to examine the error as a function of computation time, as illustrated in Figure~\ref{fig:errortime}.  
From this perspective, we conclude that DCCC consistently approximates the bounds of counterfactual queries with higher accuracy and in less time than the alternatives.

\clearpage 

\section{Conclusions}\label{sec:conclusions}


We have shown that the DCCC method, initially designed for Markovian models, can be extended to semi-Markovian models like the one in Figure~\ref{fig:modelexample}, where a confounder between the two endogenous variables is included. We have shown how to obtain a canonical representation of the domain of the exogenous variable and how the extreme points of the corresponding credal set can be identified from observational data. We have also considered the incorporation of experimental data, so that the credal set is further restricted and values of the exogenous variable that remained indistinguishable under observational data can now be discerned. Complete experimental data makes all interventional queries identifiable, whilst counterfactual queries can potentially be bounded more accurately given the information provided by the experimental data besides the observational one.

We have also proposed two ways of providing approximate answers to causal queries in semi-Markovian models by transforming them into Markovian models and then applying DCCC over the transformed one. One of the transformations consists of splitting the confounder exogenous variable, and the other one is based on merging the endogenous variables, so that a Markovian topology is reached in both cases.

We have experimentally tested that combining observational and experimental data results in tighter bounds for the causal queries. Regarding the two proposed approximations with observational data, the experiments indicate that merging the endogenous variables leads to more accurate results. Furthermore, MM-O always returns intervals that contain the true semi-Markovian interval, unlike MS-O.

Furthermore, the experiments also show how DCCC outperforms state-of-the-art methods in terms of computing time. In fact, DCCC is able to explore a higher number of solutions in a much lower time.

The methodological discussion and experimental analysis in this paper rely on a given graph structure. As future work, we plan to extend the analysis to general semi-Markovian models. \ref{sec:generalsemimarkoviandccc} contains some guidelines towards that goal.

\section*{Acknowledgments}
ARB and HL acknowledge the support by the Norwegian
Research Council through grant 304843.
RC and AS acknowledge Grant PID2022-139293NB-C31 funded by 
MICIU/AEI/10.13039/501100011033 and by ERDF ``A way of making Europe''. RC and AS acknowledge the University of Almería Research and Transfer Programme funded by
“Consejería de Universidad, Investigación e Innovación de la Junta de Andalucía”
through the European Regional Development Fund (ERDF), Operation Programme
2021-2027. Programme: Research and Innovation 54.A.
RC was also supported by ``Plan Propio de Investigaci\'{o}n y Transferencia 2024-2025''  from University of Almer\'{i}a under the project P\_LANZ\_2024/003.

\appendix
\section{Towards general semi-Markovian DCCC}
\label{sec:generalsemimarkoviandccc}

The approaches for query bounding for semi-Markovian models presented in this paper are discussed primarily for the model shown in Figure \ref{fig:modelexample}. The implementations of semi-Markovian DCCC model search employed for the experiments make use of explicit representation of the linear system to be solved, exhaustively searching through all possible extreme point solutions. 

Markvoian DCCC is on the other hand applicable generally, where for each exogenous $U$ with endogenous child $Y$, the only input required is $|\Omega_Y|$, $|\Omega_{\bmX}|$ and $\tP(Y|\bmX)$ (where $\bmX$ is the set of endogenous parents of $Y$). The structure of canonical Markovian exogenous domains allows for implicit representation of the linear system, which in addition to providing a general solver, allows for heuristic techniques to speed up the search significantly (see Section \ref{sec:solutionseach_MDCCC} and \cite{bjoru2025ijar}).

Section \ref{sec:markovianapprox_general} presents a generalisation of the Markovian endogenous merge approximation discussed in Section \ref{sec:endogenousmerge}. Based on this, Section \ref{sec:semimarkovian_general} presents a set of steps to implement model-specific semi-Markovian DCCC via Markovian DCCC, to benefit from the speed up provided by the heuristic implementation of the latter as model complexity grows.

\subsection{General Markovian approximation with endogenous merge}
\label{sec:markovianapprox_general}

Given an exogenous $U$ in a semi-Markovian SCM, let $\bmY$ be the set of all endogenous children of $U$. Let $\boldsymbol{Z}$ be the set of all endogenous variables that are both an ancestor and a descendant of variables that are in $\bmY$, along with endogenous variables that share a common exogenous parent with such a variable. Let $\bmX$ be the set of endogenous parents of variables in $\bmY\cup\boldsymbol{Z}$ not in $\bmY\cup\boldsymbol{Z}$. Let $\bmU$ be the set of exogenous parents of $\bmY\cup\boldsymbol{Z}$.

To generalise the Markovian approximation with endogenous merge presented in Section \ref{sec:markovianapprox}, consider the variables in $\bmY\cup\boldsymbol{Z}$ a single variable, with domain $\Omega_{\bmY\cup\boldsymbol{Z}}$ the Cartesian product of original variable domains. Let new exogenous parent $U^*$ replace $\bmU$. The canonical specification of $\Omega_{U^*}$ has size $|\Omega_{U^*}| = |\Omega_{\bmY\cup \boldsymbol{Z}}|^{|\Omega_\bmX|}$. The Markovian DCCC search then retrieves extreme distributions $P(U^*)$ under observed $P(\bmY\cup\boldsymbol{Z} | \bmX)$.

\subsection{General Semi-Markovian models}
\label{sec:semimarkovian_general}

For an exogenous variable $U$ in semi-Markovian models, the set of equations given by credal set $\setFont{K}(U)$ is no longer seen to correspond with the standard mapping exploited by the Markovian approach ($u_i$ is not in general represented by distinct base-$q$ encoding of $i$), and linear systems for semi-Markovian models are not solvable by direct application of DCCC from \cite{bjoru2025ijar}. However, considering an appropriate Markovian approximation, extreme solutions can be found by existing search strategies and then mapped to corresponding sets of extreme solutions in the original semi-Markovian model.
Consider first exogenous variable $U$ in the model in Figure \ref{fig:modelexample}, and its Markovian counterpart $U^*$ in the Markovian approximation seen in Figure \ref{fig:modelexample_markov}. The respective matrices of Tables \ref{tab:semimarkovianobsmatrix} and \ref{tab:markovianapprox} suggests the following steps to translate extreme points between the domains of $U$ and $U^*$:

\begin{itemize}
        \item[1.] Identify which functions indexed by the general definition domain $\Omega_{U^*}$ are not present in $\Omega_U$, and ensure these are excluded by always assigning 0 probability. For the example model, this is done by fixing $P(u^*_1) = P(u^*_4) = P(u^*_{11}) = P(u^*_{14}) = 0 $.
        \item[2.] Identify the values in $\Omega_U$ that have identical column representation by observation. In the example, this is the case for each of the sets of values $\{u_0, u_1\}$, $\{u_2, u_3\}$, $\{u_{12}, u_{14}\}$, $\{u_{13}, u_{15}\}$. There will be a single corresponding value in $\Omega_{U^*}$ (e.g. $u^*_0 \leftrightarrow $ $\{u_0, u_1\}$), such that for a single extreme point in the Markovian domain, several will be generated in the semi-Markovian domain: Each semi-Markovian extreme point assigns one of the $u_i$'s the probability of corresponding $u^*_j$, while ensuring the other $U$ values corresponding to $u^*_j$ take probability 0.
    \end{itemize}   

These steps generalise to arbitrary semi-Markovian models. With variable sets $\bmY, \boldsymbol{Z}, \bmX, \bmU$ as defined in  \ref{sec:markovianapprox_general}, and a Markovian approximation where $U^*$ is the single exogenous parent of merged endogenous child $\bmY\cup\boldsymbol{Z}$, the steps now become:

\begin{itemize}
    \item[1.] Given $\Omega_{\bmU}$, identify for which $u^*_i$ there is no counterpart in $\Omega_{\bmU}$, such that $P(u^*_i)=0$. Under these restrictions, solve for extreme $P(U^*)$ with Markovian DCCC.
    \item[2.] Identify the values in $\Omega_{\bmU}$ that have identical column representation by observation, and generate the set of extreme $P(\bmU)$ corresponding to each $P(U^*)$. If $|\bmU| > 1$, marginalise for extreme $P(U)$. 
\end{itemize}

This approach as detailed above is applicable to semi-Markovian models where only observational data is available. Note that the approach still requires the explicit structure of the domain $\Omega_\bmU$ under observed data for extreme point generation via the canonical domain $\Omega_{U^*}$ of $U^*$. While heuristic search with Markovian DCCC enable approximating query intervals for models of increased complexity compared to the example in Figure \ref{fig:modelexample}, the approach is still limited to models with endogenous domains such that the explicit exogenous domain of size $|\Omega_\bmU| = |\Omega_{\bmY\cup\boldsymbol{Z}}|^{|\Omega_{\bmX}|}$ is manageable. See the experiment section of \cite{bjoru2025ijar} for details on how the interval approximation scales as domain sizes grow.

\bibliographystyle{elsarticle-num} 
\bibliography{references}

\end{document}